\documentclass{article} 
\usepackage{iclr2027_conference,times}

\usepackage{amsmath,amsfonts,bm}

\def\eqref#1{equation~\ref{#1}}

\def\1{\bm{1}}

\DeclareMathAlphabet{\mathsfit}{\encodingdefault}{\sfdefault}{m}{sl}
\SetMathAlphabet{\mathsfit}{bold}{\encodingdefault}{\sfdefault}{bx}{n}

\usepackage{hyperref}
\usepackage{url}
\usepackage{booktabs}
\usepackage{graphicx}
\usepackage[table]{xcolor}
\usepackage{subcaption}
\usepackage{xcolor}
\usepackage{pifont}
\usepackage{multirow}
\usepackage{tabularx}
\usepackage{array}
\usepackage{amsmath}
\usepackage{amssymb}
\usepackage{algorithm}
\usepackage{algpseudocode}
\usepackage{enumitem}
\newcolumntype{Y}{>{\centering\arraybackslash}X}
\newcommand{\xmark}{\ding{55}}

\definecolor{tableblue}{RGB}{225,235,244}
\title{AirGroundVLN: A Large-Scale Benchmark for Goal-Oriented Air-Ground Collaborative Vision-and-Language Navigation}

\author{
\begin{minipage}{\dimexpr\textwidth-2\tabcolsep\relax}
\centering
\normalfont
\textbf{Zhenxuan Zeng \quad
Qingle Wu \quad
Wei Suo\setcounter{footnote}{1}\thanks{Corresponding author.} \quad
Maojia Wu \\[3pt]
Bairong Zhang} \quad
\textbf{Hangzheng Yu \quad
Peng Wang} \\[3pt]
School of Computer Science, Northwestern Polytechnical University, China.
\\[3pt]
\url{https://peakpang.github.io/AirGroundVLN-Project/}
\end{minipage}
}

\iclrfinalcopy 
\begin{document}

\maketitle
\begin{figure*}[h]
    \centering
    \includegraphics[width=\textwidth]{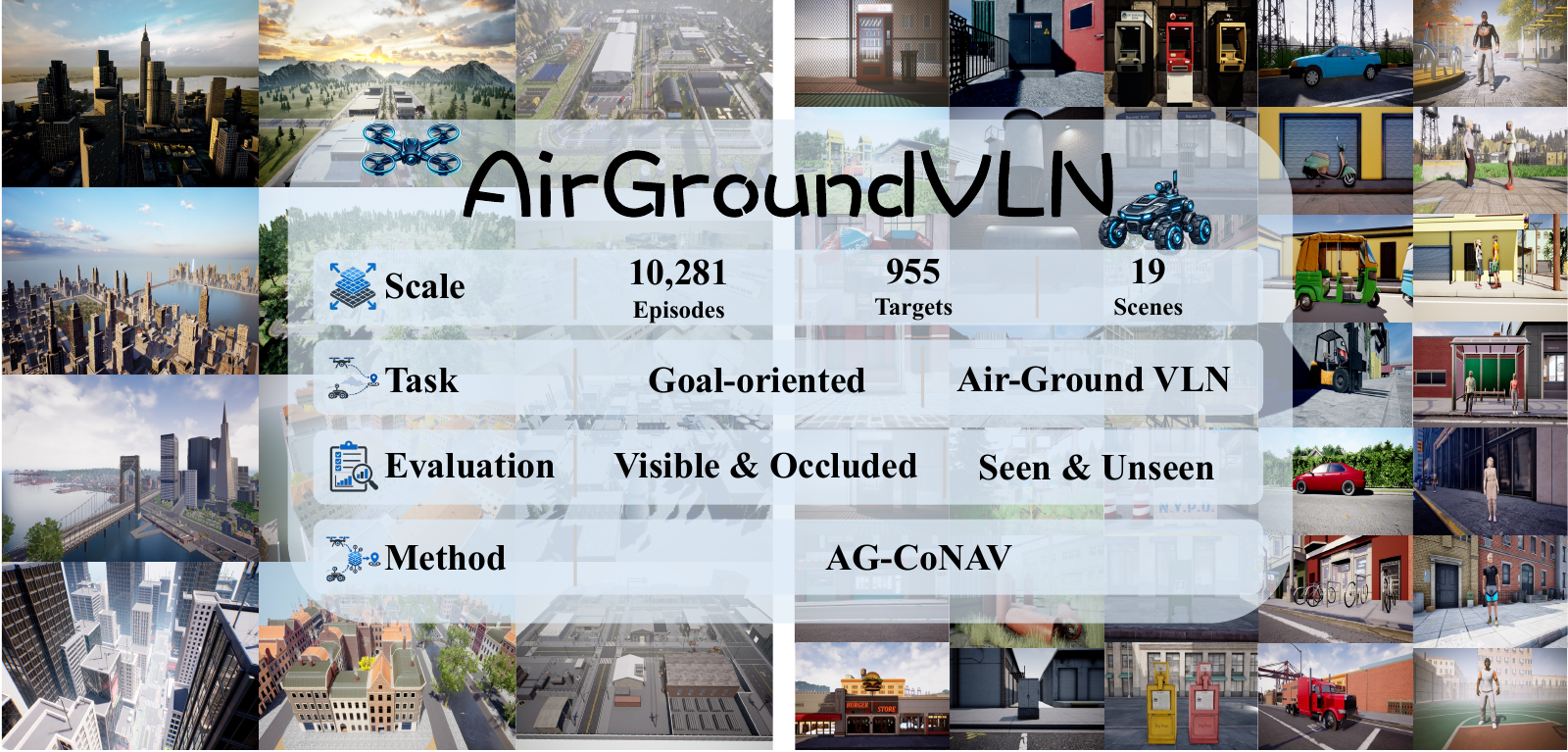}
    \caption{
\textbf{Overview of AirGroundVLN}. The benchmark contains 10,281 navigation episodes, 955 target instances, and 19 diverse scenes for goal-oriented air--ground collaborative VLN. It supports evaluation across aerial-visible/aerial-occluded target conditions and seen/unseen environments. AG-CoNAV is provided as a trainable reference framework for this setting.}
    \label{fig:benchmark}
\end{figure*}

\begin{abstract}
Goal-oriented Vision-and-Language Navigation (VLN) requires agents to locate and reach targets described in natural language without prescribed routes. Air--ground collaboration is valuable for tasks requiring both wide-area search and fine-grained localization. However, systematic study of goal-oriented air--ground collaborative VLN remains limited by the lack of large-scale, diverse benchmarks and two core challenges: 1) substantial differences between aerial and ground views, together with useful observations becoming unavailable as navigation proceeds, make it difficult to maintain spatially consistent context across platforms and over time; and 2) asymmetric spatial observability makes ground perception locally detailed but spatially limited and aerial perception broad but locally coarse, limiting the reliability of single-platform planning. To address these limitations, we introduce \textbf{AirGroundVLN}, a benchmark containing \textbf{\textit{10,281}} navigation episodes and \textbf{\textit{955}} target instances across \textbf{\textit{19}} Unreal Engine environments, with \textbf{\textit{seen/unseen}} splits and an \textbf{\textit{aerial-visibility}} protocol for systematic evaluation. Alongside the benchmark, we propose AG-CoNAV, a trainable reference framework comprising two key components: Spatiotemporally Anchored Collaborative Memory (SACM) and Aerial-Guided Regional-to-Local Planning (AGRLP). SACM maintains and retrieves spatially consistent historical context across aerial and ground observations. Meanwhile, AGRLP combines regional aerial guidance with fine-grained ground navigation. Extensive experiments demonstrate the effectiveness of AG-CoNAV and establish AirGroundVLN as a comprehensive benchmark for future exploration.
\end{abstract}

\section{Introduction}
\label{sec:intro}
Goal-oriented vision-and-language navigation (VLN) is a core problem in embodied intelligence~\citep{REVERIE,SOON}, requiring an agent to interpret a language-specified target, search for it, identify the correct instance, and reach it without prescribed intermediate routes. Existing goal-oriented VLN studies predominantly consider two types of embodied platforms: (1) \textbf{Ground Platforms}, such as wheeled Unmanned Ground Vehicles (UGVs), quadruped robots, and humanoid robots~\citep{VLN-CE,li2025ground}. (2) \textbf{Aerial Platforms}, typically Unmanned Aerial Vehicles (UAVs)~\citep{AerialVLN,TravelUAV,UAVON,Openfly}. In large-scale outdoor environments, ground platforms provide fine-grained observations of local scene structure and target appearance and can physically approach ground-level targets. In contrast, aerial platforms provide broader spatial coverage and global scene context, enabling efficient wide-area search and route-level reasoning. As two popular embodied platforms, substantial research has focused on separately expanding their perception~\citep{Vlmap}, reasoning and planning~\citep{Irpan2022, Llm-planner}, and execution capabilities~\citep{Ego,RMA}.

But consider such a real mission in a large industrial complex, where robots are instructed to \textbf{\textit{rescue a worker in a red jacket near a damaged warehouse}}. In this mission, an aerial platform can rapidly survey the area and search for the worker from a global viewpoint, but its observations may be obstructed by trees, roofs, or collapsed structures, and it cannot provide physical rescue. Meanwhile, a ground platform can closely approach the worker, inspect regions hidden from the aerial viewpoint, and provide assistance, but damaged roads, debris, and collapsed buildings may render routes infeasible. Such a mission calls for aerial and ground platforms to work together, combining wide-area search with fine-grained target localization and physical access. This complementary division of capabilities motivates heterogeneous air--ground collaboration for goal-oriented navigation.

However, systematic study of goal-oriented air--ground collaborative VLN remains limited by three key issues:  (1) Although recent efforts have explored air--ground VLN~\citep{AirGroundBench,AGC-VLN}, they remain limited in scale and diversity, with AirGroundBench and AGC-VLN evaluated on only 115 and 100 episodes, respectively. To our knowledge, a large-scale benchmark for goal-oriented air--ground collaborative VLN is still lacking, limiting systematic evaluation. (2) Beyond benchmark limitation, aerial and ground observations differ in viewpoint, while useful observations from either platform may become unavailable during navigation. Therefore, effective collaboration requires spatially consistent context across views and time. (3) Aerial and ground platforms exhibit asymmetric spatial observability: ground perception is locally detailed but spatially limited, whereas aerial perception covers broader regions but provides less fine-grained obstacle information, making either platform alone insufficient for reliable navigation.

To address this gap, we first introduce \textbf{AirGroundVLN}, a large-scale benchmark for goal-oriented air--ground collaborative VLN. AirGroundVLN contains \textbf{10,281} navigation episodes, \textbf{955} target instances, and \textbf{19} Unreal Engine environments, covering urban areas, towns, industrial areas, and other complex outdoor environments. Each episode includes aerial and ground observations together with a goal-oriented instruction describing the target appearance and its surrounding environmental context, rather than prescribing intermediate navigation actions. We provide seen and unseen environment splits to evaluate cross-environment generalization. An \textbf{\emph{aerial-visibility protocol}} is further defined to distinguish targets visible from above and those occluded by overhead structures, enabling evaluation under different target-visibility conditions.

Alongside the benchmark, we develop \textbf{AG-CoNAV}, a trainable reference framework addressing the above two key challenges of air--ground collaborative VLN. First, to maintain spatially consistent spatiotemporal context across heterogeneous aerial and ground observations, we introduce the Spatiotemporally Anchored Collaborative Memory (SACM), which stores and retrieves historical evidence by associating observations of physically related regions over time. Second, to overcome limitations of single-platform planning under asymmetric spatial observability, we introduce the Aerial-Guided Regional-to-Local Planning module (AGRLP), which combines regional aerial guidance with locally executable ground planning. In addition, to establish reference baselines for AirGroundVLN, we construct training-free and trainable methods for air--ground collaborative VLN and evaluate them alongside AG-CoNAV across seen/unseen environments and aerial-visible/aerial-occluded conditions. These evaluations establish reference performance for AirGroundVLN and demonstrate the effectiveness of AG-CoNAV. Our main contributions are summarized as follows:
\begin{itemize}

\item We introduce \textbf{AirGroundVLN}, a large-scale benchmark for goal-oriented air--ground collaborative Vision-and-Language Navigation. It contains \textbf{10,281} navigation episodes and \textbf{955} target instances across \textbf{19} Unreal Engine environments, together with goal-oriented instructions, seen/unseen splits, and an \textit{aerial-visibility} protocol for systematic evaluation.

\item We develop \textbf{AG-CoNAV}, a trainable reference framework for heterogeneous air--ground collaborative VLN. It incorporates a Spatiotemporally Anchored Collaborative Memory (SACM) for spatiotemporally consistent context association and historical retrieval, together with an Aerial-Guided Regional-to-Local Planning module (AGRLP) that coordinates regional aerial guidance with fine-grained ground execution.

\item We provide a diverse suite of training-free and trainable baselines for AirGroundVLN, establishing a comprehensive evaluation suite for this new setting. Systematic evaluations across environment generalization and aerial-visibility conditions provide reference performance for the benchmark and validate the effectiveness of AG-CoNAV.
\end{itemize}

\section{Related Work}
\label{sec:relatedwork}
\subsection{Vision-and-Language Navigation Benchmarks}
\label{subsec:rw_vln}
Early VLN benchmarks such as R2R and R4R focus on route following in indoor environments~\citep{R2R,R4R}, while Touchdown and Talk2Nav extend language-guided navigation to outdoor scenes~\citep{Touchdown,talk2nav}. VLN-CE moves navigation from discrete panoramic graphs to continuous environments~\citep{VLN-CE}, and AerialVLN further extends route-oriented VLN to UAV navigation in large-scale outdoor environments~\citep{AerialVLN}. Beyond route following, REVERIE and SOON study goal-oriented navigation toward language-described targets~\citep{REVERIE,SOON}, while TravelUAV, UAV-ON, and CityNav extend goal-oriented language guidance to aerial platforms~\citep{TravelUAV,UAVON,citynav}. Despite these advances, most benchmarks still focus on a single embodied platform. AirGroundBench begins to study heterogeneous UAV--UGV perception through 62K diagnostic VQA instances and 115 closed-loop VLN episodes~\citep{AirGroundBench}, but large-scale goal-oriented air--ground collaborative VLN remains underexplored.

\subsection{Multi-Agent Collaborative Navigation}
\label{subsec:rw_multi_agent}
Multi-agent navigation studies focus on homogeneous teams of ground robots~\citep{Burgard2000,MMM,MADE-Net} or aerial agents~\citep{MultiUAVPlanning}. Vision-language research extends collaboration to language-conditioned navigation, with CoNavBench studying VLN among ground robots~\citep{CoNavBench}, and AeroDuo and CoNav-UAV exploring complementary viewpoints within multi-UAV teams~\citep{AeroDuo,CoNav-UAV}. Moving toward heterogeneous collaboration, AGOS-Bench studies closed-loop UAV--UGV cooperative object search with visual-reference goals~\citep{AGOSBench}, while AGC-VLN introduces language-conditioned air--ground navigation through a shared bird's-eye map linking aerial context with ground planning~\citep{AGC-VLN}. However, dedicated trainable air--ground collaborative VLN remains underexplored.

\section{AirGroundVLN Benchmark}
\label{sec:airgroundvln}
In this section, we first define the goal-oriented air--ground collaborative VLN task and its evaluation protocol (Sec.~\ref{sec:task_definition}), then present the construction pipeline of AirGroundVLN (Sec.~\ref{sec:benchmark_construction}), and finally analyze its scale, diversity, and evaluation splits (Sec.~\ref{sec:dataset_analysis}).
\subsection{Task Definition}
\label{sec:task_definition}
\textbf{Task Formulation.}
AirGroundVLN considers a closed-loop navigation episode in which an Unmanned Ground Vehicle (UGV) and an Unmanned Aerial Vehicle (UAV) cooperate to locate and reach a language-specified target instance in a large-scale outdoor environment. The UGV serves as the primary target-reaching platform, while the UAV provides complementary aerial perception and global spatial context. At the beginning of each episode, the two platforms are initialized at poses $P_0^g$ and $P_0^a$, respectively, and receive a goal-oriented instruction $I$. The instruction describes the target through its semantic category, visual attributes, surrounding environmental context, and coarse relative direction, without prescribing intermediate navigation actions. At navigation step $t$, the UGV and UAV receive their current visual observations $o_t^g$ and $o_t^a$, together with historical context available at the current step. A collaborative navigation system $\pi$ produces platform-specific navigation decisions (\textit{e.g.}, waypoints or actions) and a termination signal:
\begin{equation}
(u_t^g, u_t^a, q_t)
=
\pi\left(I, o_t^g, o_t^a, H_t\right),
\qquad
q_t \in \{0,1\},
\label{eq:task_formulation}
\end{equation}
where $H_t$ denotes the historical information maintained by the navigation system at step $t$. For AirGroundVLN, the variables $u_t^g$ and $u_t^a$ denote the predicted waypoint decisions for the UGV and UAV, respectively, and $q_t=1$ indicates termination of the episode. An episode terminates at $T_{\mathrm{end}} = \min \{ t \mid q_t = 1 \text{ or } t = T_{\max} \}$, where $T_{\max}$ is the maximum allowable navigation budget.

\textbf{Evaluation Protocol.}
Given these asymmetric roles, we evaluate performance on the UGV using four standard VLN metrics~\citep{anderson2018evaluation}: Success Rate (SR), Success weighted by Path Length (SPL), UGV Navigation Error (NE\textsubscript{UGV}), and Oracle Success Rate (OSR). For the UAV, we report UAV Navigation Error (NE\textsubscript{UAV}), defined as its final horizontal distance to the target, as a \textit{diagnostic} metric. Evaluation protocol and metric definitions are provided in Appendix~\ref{supp:metrics}.

\subsection{Benchmark Construction}
\label{sec:benchmark_construction}
\begin{figure*}[h]
    \centering
    \includegraphics[width=\textwidth]{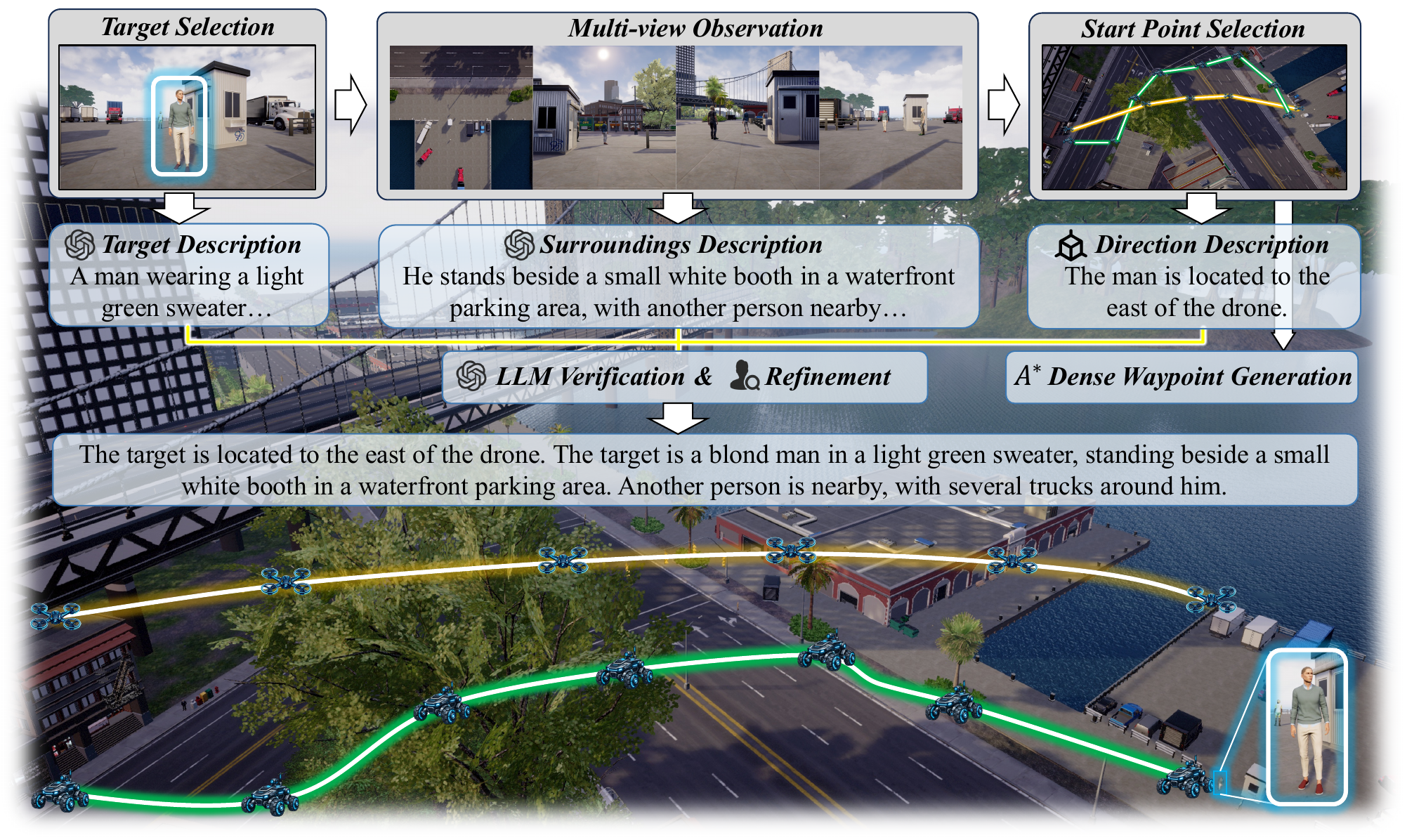}
    \caption{
    \textbf{Overview of AirGroundVLN data collection and an illustrative episode}. The collection process includes target and environment description, start-location selection, goal-oriented instruction generation, and dense air--ground waypoint generation.
    }
    \label{fig:data_collect_pipeline}
\end{figure*}

AirGroundVLN comprises 19 outdoor scenes in Unreal Engine, covering urban areas, towns, industrial areas, and other complex outdoor environments. The simulation environment is built in Unreal Engine using AirSim~\citep{Airsim} interfaces for vehicle control and sensing, with CARLA-Air interface compatibility~\citep{CARLA-Air}. The benchmark construction pipeline consists of four components:
\textbf{(1) Goal-Oriented Instruction Generation.}
As illustrated in Fig.~\ref{fig:data_collect_pipeline}, for each target selected by annotators, we collect multi-view observations to describe its appearance and surrounding environment, and derive a coarse relative direction for each start location.
\textbf{(2) Air--Ground Trajectory Generation.}
Following prior aerial VLN benchmark construction~\citep{Openfly}, we generate platform-specific reference trajectories using A$^*$-based planning~\citep{Astar}. For the UGV, annotators specify sparse key waypoints that respect road topology and traversability, between which a modified surface-based A$^*$ planner generates the ground trajectory. UAV reference trajectories are generated directly using A$^*$ from the start pose to the end pose.
\textbf{(3) Visibility Annotation.}
In goal-oriented air--ground collaborative navigation, target instances may be occluded from the aerial viewpoint by overhead structures. To systematically evaluate collaborative navigation under different aerial target-visibility conditions, we categorize target instances into \textbf{\textit{aerial-visible}} and \textbf{\textit{aerial-occluded}} subsets.
\textbf{(4) Human Quality Assurance.}
Human verification is incorporated throughout construction. Notably, all candidate episodes are checked for consistency across target instances, instructions, air--ground trajectories, and aerial-visibility labels. Recoverable inconsistencies are manually corrected, while ambiguous or invalid samples are discarded, followed by a final comprehensive review to obtain the resulting benchmark. Detailed construction procedures and quality-control statistics are provided in Appendix~\ref{supp:benchmark_construction}.

\subsection{Dataset Analysis}
\label{sec:dataset_analysis}
\begin{table*}[h]
    \centering
    \caption{Comparison with representative VLN benchmarks. $N_{\mathrm{traj}}$ denotes the number of navigation episodes (paired air--ground trajectories for AirGroundVLN), while $L_{\mathrm{traj}}$ and $L_{\mathrm{instr}}$ denote average trajectory length (m) and instruction length (words), respectively. ``--'' indicates statistics not officially reported by the corresponding benchmark.}
    \label{tab:dataset_comparison}
    \scriptsize
    \setlength{\tabcolsep}{3.5pt}
    \renewcommand{\arraystretch}{1.12}
    \resizebox{0.9\textwidth}{!}{
    \begin{tabular}{lcccccccc}
        \hline
        \textbf{Benchmark} & \textbf{Ground} & \textbf{Air} & \textbf{Simulator} & $\mathbf{N_{\mathrm{traj}}}$ & $\mathbf{L_{\mathrm{traj}}^{\mathrm{ground}}}$ & $\mathbf{L_{\mathrm{traj}}^{\mathrm{aerial}}}$ & $\mathbf{L_{\mathrm{instr}}}$ & $\mathbf{N_{\mathrm{vocab}}}$ \\
        \hline
        R2R~\citep{R2R} & $\surd$ & $\times$ & Matterport3D & 7,189 & 10.0 & $\times$ & 29 & 3.1K \\
        REVERIE~\citep{REVERIE} & $\surd$ & $\times$ & Matterport3D & 7,234 & 10.0 & $\times$ & 18 & 1.6K \\
        RxR~\citep{RR} & $\surd$ & $\times$ & Matterport3D & 13,992 & 14.9 & $\times$ & 129 & 7.0K \\
        VLN-CE~\citep{VLN-CE} & $\surd$ & $\times$ & Habitat & 4,475 & 11.1 & $\times$ & -- & 4.3K \\
        \cmidrule(lr){1-9}
        AerialVLN~\citep{AerialVLN} & $\times$ & $\surd$ & AirSim & 8,446 & $\times$ & 661.8 & 83 & 4.5K \\
        TravelUAV~\citep{TravelUAV} & $\times$ & $\surd$ & OpenUAV & 12,149 & $\times$ & 255.0 & 104 & 10.8K \\
        CityNav~\citep{citynav} & $\times$ & $\surd$ & CityFlight & 32,637 & $\times$ & 545.0 & 26 & 6.4K \\
        UAV-ON~\citep{UAVON} & $\times$ & $\surd$ & AirSim & 11,000 & $\times$ & -- & -- & -- \\
        HaL-13k~\citep{AeroDuo} & $\times$ & $\surd$ & OpenUAV & 13,838 & $\times$ & 208.4 & -- & -- \\
        \cmidrule(lr){1-9}
        AirGroundBench~\citep{AirGroundBench} & $\surd$ & $\surd$ & AirSim & 115 & -- & -- & -- & -- \\
        AGC-VLN~\citep{AGC-VLN} & $\surd$ & $\surd$ & CARLA-Air & 100 & -- & -- & -- & -- \\
        \textbf{AirGroundVLN (Ours)} & $\mathbf{\surd}$ & $\mathbf{\surd}$ & AirSim/CARLA-Air & \textbf{10,281} & \textbf{90.9} & \textbf{79.0} & \textbf{82} & \textbf{2.5K} \\
        \hline
    \end{tabular}}
\end{table*}

\begin{figure*}[h]
    \centering
    \includegraphics[width=\textwidth]{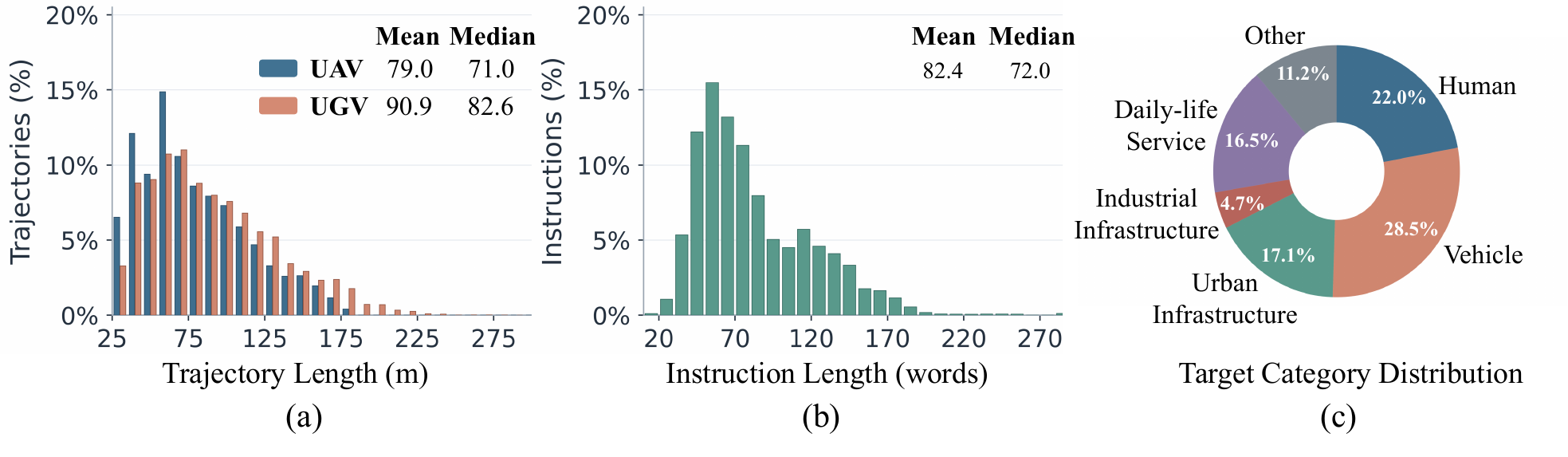}
    \caption{
    \textbf{Statistics of AirGroundVLN}. (a) Distribution of UAV and UGV reference trajectory lengths. (b) Distribution of navigation instruction lengths. (c) Distribution of target instances across six major semantic categories.
    }
    \label{fig:dataset_statistics}
\end{figure*}

As summarized in Table~\ref{tab:dataset_comparison}, AirGroundVLN extends existing VLN benchmarks to large-scale goal-oriented navigation with air–ground agents. Most prior datasets focus on single-platform ground or aerial navigation, while existing air--ground benchmarks lack scale and comprehensive evaluation. By contrast, AirGroundVLN provides $10{,}281$ air--ground collaborative navigation episodes across $19$ simulation scenes, covering $955$ target instances. This benchmark enables systematic evaluation of how heterogeneous aerial and ground observations can be jointly exploited for goal-oriented navigation. We further analyze the key characteristics of the benchmark in Fig.~\ref{fig:dataset_statistics}.

\textbf{Trajectory, Language, and Target Diversity.}
As shown in Fig.~\ref{fig:dataset_statistics} (a), the average length of UAV reference trajectories in AirGroundVLN is $79.0$ m , whereas the corresponding UGV trajectories average $90.9$ m. Ground trajectories are longer on average, reflecting the different motion constraints of the two platforms: UAV trajectories are less constrained by ground topology, whereas UGVs must follow traversable surfaces and road structures. As shown in Fig.~\ref{fig:dataset_statistics} (b), navigation instructions in AirGroundVLN average $82$ words. Furthermore, as depicted in Fig.~\ref{fig:dataset_statistics} (c), the $955$ target instances span six broad semantic categories, demonstrating substantial semantic diversity across vehicles, humans, urban infrastructure, daily-life services, industrial infrastructure, and other objects.

\textbf{Data Splits and Visibility.}
AirGroundVLN is divided into \textit{train}, \textit{val-seen}, \textit{test-seen}, and \textit{test-unseen} splits with $7{,}538$, $933$, $962$, and $848$ episodes, respectively, where \textit{test-unseen} evaluates generalization to unseen environments. To evaluate air--ground collaborative VLN under different aerial target-visibility conditions, we categorize test episodes as \textit{aerial-visible} or \textit{aerial-occluded} based on visibility from the aerial viewpoint. Further benchmark statistics are provided in Appendix~\ref{supp:data_splits} (\textit{e.g.}, word clouds of target names and visual attributes, and aerial-visibility distributions).

\section{Method}
\label{sec:method}

\begin{figure*}[h]
\centering
\includegraphics[width=\textwidth]{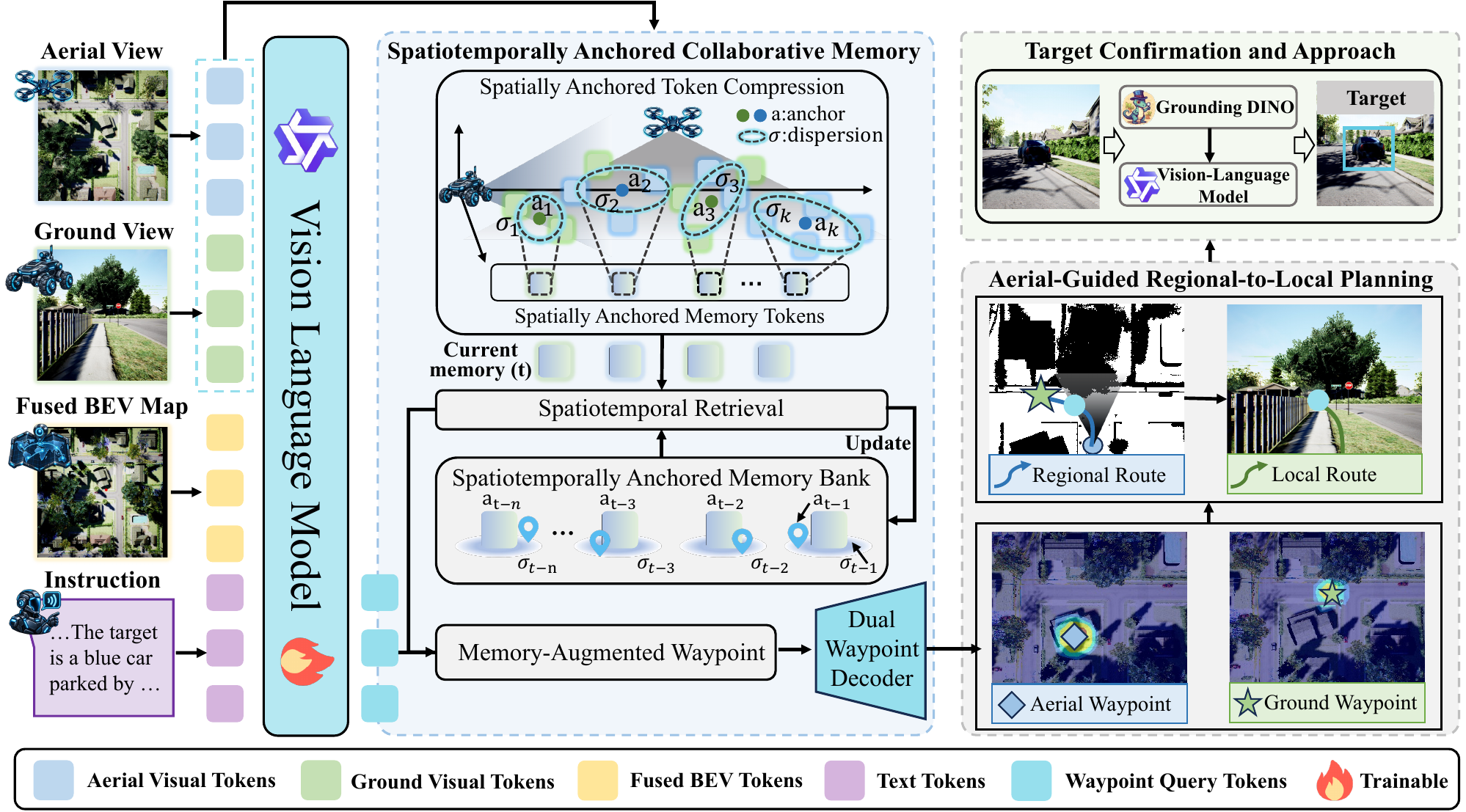}
\caption{\textbf{Overview of the AG-CoNAV framework}. Given goal-oriented instructions, aerial--ground observations, and a UAV-maintained fused BEV map, SACM maintains and retrieves spatially consistent context across views and time, while AGRLP coordinates regional aerial guidance with fine-grained ground execution.}
\label{fig:method_overview}
\end{figure*}

We propose \textbf{AG-CoNAV}, a trainable framework for goal-oriented air--ground collaborative VLN designed to tackle cross-view spatiotemporal context inconsistency and planning under asymmetric spatial observability. As illustrated in Fig.~\ref{fig:method_overview}, given a goal-oriented instruction, aerial--ground observations, and a UAV-maintained fused BEV map constructed by projecting and fusing sequential UAV RGB-D observations, the Spatiotemporally Anchored Collaborative Memory (SACM) (Sec.~\ref{subsec:sacm}) maintains and retrieves spatially consistent context across views and time. This retrieved context is then incorporated into multimodal waypoint queries to predict UAV and UGV waypoint distributions. Finally, the Aerial-Guided Regional-to-Local Planning module (AGRLP) (Sec.~\ref{subsec:agrlp}) uses these predictions to coordinate regional aerial guidance with fine-grained ground navigation.

\subsection{Spatiotemporally Anchored Collaborative Memory} 
\label{subsec:sacm}  
Aerial and ground observations offer complementary information but differ in viewpoint, scale, and spatial coverage. Moreover, evidence observed at one navigation step may become unavailable in later observations due to occlusion and changing viewpoints. Effective collaboration therefore requires preserving complementary evidence across views and time while maintaining spatial consistency. To this end, we introduce SACM, which compresses observations into spatially characterized memory tokens and retrieves historical evidence via dispersion-aware spatiotemporal matching.

\textbf{Spatially Anchored Memory Construction.}
To preserve historical evidence efficiently, we compress visual features into a compact set of memory tokens. Conventional feature-only compression discards the physical origin of the aggregated evidence. Associating each compressed token with a single coordinate partially restores this information, but cannot distinguish tokens concentrated around a local region from those aggregating evidence over a much broader area. \textit{This distinction is particularly important for air--ground collaboration, where aerial and ground observations naturally differ in spatial coverage.} We therefore characterize each compressed token with both a spatial anchor and a spatial dispersion to regulate historical retrieval. For each platform $v\in\{g,a\}$, let $X_t^v\in\mathbb{R}^{N_v\times d}$ denote its visual tokens at step $t$. Using camera parameters and depth, each visual token is projected onto the shared ground plane with metric coordinate $\mathbf{c}_{t,i}^v\in\mathbb{R}^2$. View-specific learned queries compress $X_t^v$ into $K$ memory tokens $M_t^v\in\mathbb{R}^{K\times d}$. Let $\alpha_{t,k,i}^v$ denote the normalized cross-attention weight ($\sum_i\alpha_{t,k,i}^v=1$). Each compressed token is characterized by a spatial anchor $\mathbf{a}_{t,k}^v$ and dispersion $\sigma_{t,k}^v$:
\begin{equation}
\mathbf{a}_{t,k}^v
=
\sum_i \alpha_{t,k,i}^v \mathbf{c}_{t,i}^v,
\qquad
\sigma_{t,k}^v
=
\sqrt{
\sum_i \alpha_{t,k,i}^v
\left\|
\mathbf{c}_{t,i}^v-\mathbf{a}_{t,k}^v
\right\|_2^2
}.
\label{eq:spatial_anchor}
\end{equation}
Thus, $\mathbf{a}_{t,k}^v$ captures where the compressed evidence is centered, whereas $\sigma_{t,k}^v$ captures how broadly that evidence is distributed in physical space. The resulting tokens and their metadata are stored in a shared air--ground memory bank.

\textbf{Spatiotemporal Memory Retrieval.} Historical evidence is retrieved through cross-attention augmented with spatial and temporal priors, allowing current tokens to aggregate complementary evidence across aerial and ground views. Let $\mathcal{B}_{t-1}=\{(\mathbf{m}_j,\mathbf{a}_j,\sigma_j,\tau_j)\}_j$ denote the historical memory bank accumulated from both platforms, where $\mathbf{m}_j$, $\mathbf{a}_j$, $\sigma_j$, and $\tau_j$ are the feature, spatial anchor, dispersion, and timestamp of the $j$-th memory token, respectively. For each current--historical token pair, the semantic attention score incorporates spatial and temporal priors. Let $\Delta t_j=t-\tau_j$ denote the relative age of historical token $j$. The retrieval score is
\begin{equation}
S_{ij}
=
\frac{\mathbf{q}_i^\top\mathbf{k}_j}{\sqrt{d}}
-
\lambda_{\mathrm{spa}}
\frac{\|\mathbf{a}_i-\mathbf{a}_j\|_2^2}
{\rho^2+\sigma_i^2+\sigma_j^2}
-
\lambda_{\mathrm{tmp}}\log(1+\Delta t_j),
\label{eq:retrieval_score}
\end{equation}
where $d$ is the feature dimension, $\rho$ defines a base spatial tolerance, and
$\lambda_{\mathrm{spa}}$ and $\lambda_{\mathrm{tmp}}$ are learnable weights for
the spatial and temporal priors, respectively. The dispersion-aware spatial term favors closer anchor alignment for spatially concentrated evidence, while allowing greater positional variation for tokens spanning broader regions. Thus, retrieval accounts for both where the evidence is located and how much physical space it represents. The retrieved historical evidence is aggregated into a spatiotemporally consistent context $F_t^{\mathrm{mem}}$, while the current memory tokens and their spatial metadata are appended to update the memory bank. Finally, waypoint queries from the multimodal backbone attend to $F_t^{\mathrm{mem}}$, yielding the memory-augmented waypoint queries $Q_t^{\mathrm{wp}}$ used by AGRLP.

\subsection{Aerial-Guided Regional-to-Local Planning}
\label{subsec:agrlp}
While ground observations ensure locally safe navigation, their limited spatial scope can lead to short-sighted planning. To overcome this asymmetric perception, AGRLP couples regional aerial guidance with hierarchical ground planning, leveraging broad spatial context while preserving locally executable navigation.

\textbf{Dual Waypoint Prediction.} Given the memory-augmented waypoint queries $Q_t^{\mathrm{wp}}$, following heatmap-based waypoint prediction~\citep{Hong2022Bridging}, we model aerial and ground waypoints as spatial distributions. The waypoints are obtained as $(\hat{\mathbf{p}}_t^g,\hat{\mathbf{p}}_t^a)=\Phi(Q_t^{\mathrm{wp}})$, where $\Phi$ denotes the waypoint prediction and decoding function. We jointly supervise the predicted distributions and their spatial centers, with the training loss detailed in Appendix~\ref{supp:training objective}.

\textbf{Regional-to-Local Planning and Target Approach.}
The UAV maintains a regional occupancy grid $\mathrm{Occ}_t^a$ from depth observations while exploring toward $\hat{\mathbf{p}}_t^a$. Given the current UGV position $\mathbf{x}_t^g$ and predicted ground waypoint $\hat{\mathbf{p}}_t^g$, the regional route is computed as
$P_t^{\mathrm{regional}}=\operatorname{A}^*(\mathbf{x}_t^g,\hat{\mathbf{p}}_t^g;\mathrm{Occ}_t^a)$,
providing long-range navigation guidance. Meanwhile, the UGV maintains a local occupancy grid $\mathrm{Occ}_t^g$ from its RGB-D observations, selects a reachable sub-goal along $P_t^{\mathrm{regional}}$ within the local traversable space, and computes a collision-free local trajectory toward it using A* search. Following~\citep{TravelUAV}, we detect target candidates with Grounding DINO~\citep{GroundingDINO} and verify their appearance and surrounding context against the instruction. Upon confirmation, the UGV approaches the candidate's position estimated from RGB-D observations~\citep{apex} and terminates the episode (Appendix~\ref{supp:target_approach}).
\section{Experiments}
\label{sec:experiment}

\subsection{Experimental Setup}
\label{subsec:experimental_setup}
AG-CoNAV uses Qwen3.5-2B~\citep{Qwen35} as the vision-language backbone. We freeze the pretrained backbone parameters and train LoRA adapters~\citep{LoRA}, while jointly optimizing the memory and waypoint prediction modules. Implementation details are provided in Appendix~\ref{supp:experiments}. All experiments follow the evaluation protocol in Sec.~\ref{sec:task_definition}. The success distance threshold is set to 5~m.

\subsection{Baselines}
\label{subsec:baselines}
\textbf{Random Policy.}
We include a random navigation policy as a lower-bound reference. At each step, the UGV randomly samples and follows a waypoint within its local collision-free space.

\textbf{Training-Free Baselines.}
We evaluate several representative Vision-Language Models (VLMs), including Qwen3.5-2B/4B~\citep{Qwen35}, Gemini-3.6-Flash~\citep{gemini}, and GPT-5.5~\citep{GPT55}, under a unified zero-shot air--ground navigation pipeline. The evaluated models span high-capacity proprietary VLMs and compact open-source VLMs. We also include AGC-VLN~\citep{AGC-VLN}, a training-free air--ground collaborative VLN method. Details of the unified training-free navigation pipeline are provided in Appendix~\ref{supp:experiments}.

\textbf{Training-Based Baselines.}
To examine the effects of global spatial context and heterogeneous air--ground collaboration, we construct three training-based baselines.
\textbf{(1) UGV Only}: This baseline predicts ground waypoints solely from ground observations without aerial assistance. 
\textbf{(2) UGV + Global BEV Map}: This baseline incorporates an aerially derived global BEV map into ground waypoint prediction to compare static global context with active collaboration. 
\textbf{(3) AeroDuo}: To our knowledge, no existing trainable framework is designed for heterogeneous air--ground collaborative VLN. We therefore adapt AeroDuo~\citep{AeroDuo} to the UAV--UGV setting of AirGroundVLN.

\subsection{Main Results}
\label{subsec:main_results}
\begin{table*}[h]
  \centering
  \caption{
  Comparison with baselines on AirGroundVLN.
  Results are reported across seen and unseen environments and under
  different aerial visibility conditions.
  }
  \label{tab:main_results}
  \resizebox{0.9\textwidth}{!}{%
  \begin{tabular}{@{}l*{10}{c}@{}}
  \toprule

  \textbf{Method} 
  & \multicolumn{5}{c}{\textbf{Seen}} 
  & \multicolumn{5}{c}{\textbf{Unseen}} \\
  \cmidrule(lr){2-6} 
  \cmidrule(lr){7-11}

  & \textbf{SR$\uparrow$}
  & \textbf{OSR$\uparrow$}
  & \textbf{SPL$\uparrow$}
  & \textbf{NE\textsubscript{UGV}$\downarrow$}
  & \textbf{NE\textsubscript{UAV}$\downarrow$}
  & \textbf{SR$\uparrow$}
  & \textbf{OSR$\uparrow$}
  & \textbf{SPL$\uparrow$}
  & \textbf{NE\textsubscript{UGV}$\downarrow$}
  & \textbf{NE\textsubscript{UAV}$\downarrow$} \\
  \midrule

  Random 
  & 0.52 & 0.52 & 0.34 & 79.66 & --
  & 0.47 & 0.59 & 0.23 & 72.43 & -- \\

  \midrule
  \rowcolor{tableblue}
  \multicolumn{11}{l}{\textbf{\textit{Training-Free Methods}}}\\
  
  Air-Ground (Qwen3.5-2B)
  & 4.05 & 4.26 & 2.38 & 81.08 & 100.85
  & 1.30 & 1.77 & 0.89 & 69.60 & 92.12 \\

  Air-Ground (Qwen3.5-4B)
  & 13.10& 15.59& 9.54& 63.33& 68.47
  & 10.14& 11.56& 6.75& 54.96& 62.58 \\

  AGC-VLN (Qwen3.5-4B)
  & 13.41& 16.63& 8.42& 65.75& 63.99
  & 10.50& 12.38& 4.95& 54.37& 51.16\\
  
  Air-Ground (Gemini-3.6-Flash)
  & 22.25 & 22.77& 20.51 & \underline{47.80} & \underline{50.59}
  & 17.45& 18.99 & 15.79& 46.41& 51.82\\

  Air-Ground (GPT-5.5)
  & \underline{24.43} & \underline{25.99} & \underline{21.59} & 48.77 & 54.78 
  & \underline{19.69} & \underline{22.05} & \underline{17.25} & \textbf{44.21} & \underline{48.07} \\
  \midrule
  \rowcolor{tableblue}
  \multicolumn{11}{l}{\textbf{\textit{Training-Based Methods}}}\\

  UGV Only
  & 16.84 & 20.89 & 10.86 & 62.92 & --
  & 13.33 & 17.69 & 8.38 & 57.43 & -- \\

  UGV + Global BEV Map
  & 20.06 & 24.53 & 18.81 & 53.13 & --
  & 16.86 & 20.28 & 15.66 & 45.21 & -- \\

  AeroDuo
  & 17.46 & 22.45 & 14.00 & 54.71 & 56.27
  & 15.45 & 19.69 & 13.79 & 46.17 & 55.88 \\

  \textbf{AG-CoNAV (Qwen3.5-2B)}
  & \textbf{32.64} & \textbf{35.65} & \textbf{29.85} & \textbf{39.72} & \textbf{44.81}
  & \textbf{20.05} & \textbf{24.06} & \textbf{18.46} & \underline{44.43} & \textbf{44.04} \\

  \midrule[1.5pt]

  \textbf{Method} 
  & \multicolumn{5}{c}{\textbf{Visible}} 
  & \multicolumn{5}{c}{\textbf{Occluded}} \\
  \cmidrule(lr){2-6} 
  \cmidrule(lr){7-11}

  & \textbf{SR$\uparrow$}
  & \textbf{OSR$\uparrow$}
  & \textbf{SPL$\uparrow$}
  & \textbf{NE\textsubscript{UGV}$\downarrow$}
  & \textbf{NE\textsubscript{UAV}$\downarrow$}
  & \textbf{SR$\uparrow$}
  & \textbf{OSR$\uparrow$}
  & \textbf{SPL$\uparrow$}
  & \textbf{NE\textsubscript{UGV}$\downarrow$}
  & \textbf{NE\textsubscript{UAV}$\downarrow$} \\
  \midrule

  Random
  & 0.64 & 0.71 & 0.37 & 78.41 & -- 
  & 0.00 & 0.00 & 0.00 & 68.96 & -- \\

  \midrule
  \rowcolor{tableblue}
  \multicolumn{11}{l}{\textbf{\textit{Training-Free Methods}}}\\
  
  Air-Ground (Qwen3.5-2B)
  & 3.14 & 3.43 & 1.93 & 77.83 & 96.77
  & 1.46 & 1.95 & 0.83 & 68.41 & 96.71 \\
  
  Air-Ground (Qwen3.5-4B)
  & 12.57& 14.57& 9.00& 61.29& 67.79
  & 8.78& 10.73& 5.60& 52.96& 58.60 \\

  AGC-VLN (Qwen3.5-4B)
  & 12.86& 14.86& 6.88& 61.97& 61.88
  & 9.27& 13.90& 6.50& 55.10& \underline{44.64}\\
  
  Air-Ground (Gemini-3.6-Flash)
  &21.43 &22.36 &19.76 & \underline{46.67}& \underline{51.42}
  &15.12 &16.34 &13.32 & 48.78& 50.33 \\

  Air-Ground (GPT-5.5)
  & \underline{23.29} & \underline{24.86} & \underline{20.58} & 47.38 & 53.12
  & \underline{18.54} & \underline{21.71} & \underline{16.05} & \underline{44.09} & 46.60 \\
  \midrule
  \rowcolor{tableblue}
  \multicolumn{11}{l}{\textbf{\textit{Training-Based Methods}}}\\

  UGV Only
  & 16.36 & 20.64 & 10.07 & 61.35 & --
  & 11.22 & 15.12 & 8.42 & 56.90 & -- \\

  UGV + Global BEV Map
  & 19.64 & 24.36 & 18.89 & 49.73 & --
  & 14.88 & 16.34 & 12.03 & 48.35 & -- \\

  AeroDuo
  & 17.36 & 22.71 & 14.84 & 51.29 & 56.64
  & 13.66 & 15.85 & 10.68 & 48.75 & 54.21 \\

  \textbf{AG-CoNAV (Qwen3.5-2B)}
  & \textbf{28.86} & \textbf{31.86} & \textbf{26.54} & \textbf{42.12} & \textbf{45.21}
  & \textbf{19.51} & \textbf{24.63} & \textbf{17.59} & \textbf{41.25} & \textbf{41.84} \\
  \bottomrule
  \end{tabular}
   }
\end{table*}

\textbf{Performance Analysis.}
Table~\ref{tab:main_results} compares training-free and training-based baselines on AirGroundVLN. AG-CoNAV achieves the best SR across all splits, demonstrating its effectiveness for air--ground collaborative VLN. Compared with the training-free Air-Ground model using the same Qwen3.5-2B backbone, AG-CoNAV substantially improves SR from 4.05\% to 32.64\% and SPL from 2.38\% to 29.85\% on the seen split, highlighting the advantage of task-specific training. Compared with \textit{UGV + Global BEV Map}, AG-CoNAV consistently achieves higher success rates under both visible and occluded conditions, demonstrating the benefit of actively acquired aerial context over static global information. Further \textit{\textbf{qualitative trajectory comparisons}} are provided in Appendix~\ref{supp:qualitative}. An animated \textit{\textbf{simulation demo of AG-CoNAV}} showcasing the closed-loop air--ground navigation trajectories is provided in the supplementary material.

\subsection{Ablation Studies}
\label{subsec:ablation}
In this section, we ablate SACM and AGRLP to evaluate the effectiveness of each component in AG-CoNAV.

\textbf{Influence of Main Modules.}
We first evaluate the contribution of each module in AG-CoNAV. The variant
without SACM and AGRLP serves as the baseline. As shown in
Table~\ref{tab:module_ablation}, introducing AGRLP improves the baseline SR
from 25.68\% to 26.30\% and SPL from 22.63\% to 24.03\% on the seen split,
demonstrating the benefit of aerial guidance for ground navigation. Adding SACM alone further increases the SR to 31.39\% and SPL to 27.79\%, indicating the
importance of spatially consistent historical context retrieval. Combining
SACM and AGRLP achieves the best performance, with SR and SPL reaching 32.64\%
and 29.85\%, respectively, showing that the two components provide
complementary benefits for air--ground collaborative navigation.

\textbf{Influence of SACM Priors.}
We further investigate the contribution of the temporal prior and the dispersion-aware spatial prior in SACM. As shown in Table~\ref{tab:sacm_prior_ablation}, removing either prior degrades performance compared with the complete SACM. Specifically, using only the temporal prior achieves 29.83\% SR on the seen split, while using only the spatial prior achieves 31.19\% SR. Combining both priors further increases the SR to 32.64\% and consistently improves other navigation metrics. These results demonstrate that temporal continuity and dispersion-aware spatial context provide complementary information for retrieving relevant historical observations across views and time. 

\begin{table*}[h]
\centering
\caption{
Ablation of SACM and AGRLP components in AG-CoNAV on seen/unseen environments.
}
\label{tab:module_ablation}
\setlength{\tabcolsep}{4pt}
\renewcommand{\arraystretch}{1.12}
\resizebox{0.85\textwidth}{!}{
\begin{tabular}{cc|ccccc|ccccc}
\toprule
\multirow{2}{*}{\textbf{SACM}}
& \multirow{2}{*}{\textbf{AGRLP}}
& \multicolumn{5}{c|}{\textbf{Seen}}
& \multicolumn{5}{c}{\textbf{Unseen}} \\
\cmidrule(lr){3-7}
\cmidrule(lr){8-12}

&
&
\textbf{SR$\uparrow$}
& \textbf{OSR$\uparrow$}
& \textbf{SPL$\uparrow$}
& \textbf{NE\textsubscript{UGV}$\downarrow$}
& \textbf{NE\textsubscript{UAV}$\downarrow$}
& \textbf{SR$\uparrow$}
& \textbf{OSR$\uparrow$}
& \textbf{SPL$\uparrow$}
& \textbf{NE\textsubscript{UGV}$\downarrow$}
& \textbf{NE\textsubscript{UAV}$\downarrow$} \\
\midrule

\xmark
& \xmark
& 25.68 & 27.65 & 22.63 & 48.59 & 55.21
& 17.10 & 19.93 & 15.61 & 50.02 & 49.30 \\

\xmark
& \checkmark
& 26.30 & 29.42 & 24.03 & 46.71 & 54.49
& 17.92 & 20.64 & 16.18 & 48.59 & 48.76 \\

\checkmark
& \xmark
& 31.39 & 33.47 & 27.79 & 41.93 & 45.62
& 19.22 & 22.29 & 17.56 & 45.76 & 45.13 \\

\checkmark
& \checkmark
& \textbf{32.64} & \textbf{35.65} & \textbf{29.85} & \textbf{39.72} & \textbf{44.81}
& \textbf{20.05} & \textbf{24.06} & \textbf{18.46} & \textbf{44.43} & \textbf{44.04} \\

\bottomrule
\end{tabular}}
\end{table*}

\begin{table*}[h]
\centering
\caption{
Detailed ablation of the temporal and dispersion-aware spatial priors in SACM on seen and unseen environments. AGRLP is enabled for all variants.
}
\label{tab:sacm_prior_ablation}
\setlength{\tabcolsep}{4pt}
\renewcommand{\arraystretch}{1.12}

\resizebox{0.9\textwidth}{!}{
\begin{tabular}{cc|ccccc|ccccc}
\toprule
\multirow{2}{*}{\textbf{Temporal Prior}}
& \multirow{2}{*}{\textbf{Spatial Prior}}
& \multicolumn{5}{c|}{\textbf{Seen}}
& \multicolumn{5}{c}{\textbf{Unseen}} \\
\cmidrule(lr){3-7}
\cmidrule(lr){8-12}

&
&
\textbf{SR$\uparrow$}
& \textbf{OSR$\uparrow$}
& \textbf{SPL$\uparrow$}
& \textbf{NE\textsubscript{UGV}$\downarrow$}
& \textbf{NE\textsubscript{UAV}$\downarrow$}
& \textbf{SR$\uparrow$}
& \textbf{OSR$\uparrow$}
& \textbf{SPL$\uparrow$}
& \textbf{NE\textsubscript{UGV}$\downarrow$}
& \textbf{NE\textsubscript{UAV}$\downarrow$} \\
\midrule

\checkmark
& \xmark
& 29.83 & 31.91 & 26.75 & 43.92 & 47.03
& 18.99 & 21.82 & 16.92 & 47.14 & 46.19 \\

\xmark
& \checkmark
& 31.19 & 34.10 & 28.56 & 41.37 & 45.92
& 19.69 & 23.11 & 17.53 & 45.61 & 44.86 \\

\checkmark
& \checkmark
& \textbf{32.64} & \textbf{35.65} & \textbf{29.85} & \textbf{39.72} & \textbf{44.81}
& \textbf{20.05} & \textbf{24.06} & \textbf{18.46} & \textbf{44.43} & \textbf{44.04} \\

\bottomrule
\end{tabular}}
\end{table*}

\section{Conclusion}
\label{sec:conclusion}
We introduce \textbf{AirGroundVLN}, a large-scale benchmark for goal-oriented air--ground collaborative Vision-and-Language Navigation, containing 10,281 episodes and 955 target instances across 19 Unreal Engine environments, equipped with an aerial-visibility evaluation protocol. Alongside the benchmark, we propose \textbf{AG-CoNAV}, a framework addressing heterogeneous collaboration via Spatiotemporally Anchored Collaborative Memory (SACM) and Aerial-Guided Regional-to-Local Planning (AGRLP). Extensive evaluations validate AG-CoNAV and establish strong baselines. We hope AirGroundVLN can serve as a useful testbed for the community and facilitate future research on heterogeneous air--ground collaborative VLN toward more capable embodied navigation systems.

\section*{AI Use Statement}
In this work, we used generative AI tools to assist with instruction generation during dataset construction. All generated instructions were manually reviewed and verified by the authors (Sec.~\ref{sec:benchmark_construction}). We also used generative AI tools to check grammatical errors and improve the clarity and readability of the manuscript. We take responsibility for the final content of this work.

\bibliography{iclr2027_conference}
\bibliographystyle{iclr2027_conference}

\appendix
\section*{Appendix}
\section{Evaluation Protocol and Metrics}
\label{supp:metrics}
\textbf{Metrics.} Consider an evaluation split containing $N$ episodes. For the $i$-th episode, let $\mathbf{x}_{i,t}^g \in \mathbb{R}^3$ and
$\mathbf{x}_{i,t}^a \in \mathbb{R}^3$ denote the positions of the UGV and UAV
at navigation step $t$, respectively, and let
$\mathbf{x}_i^{\mathrm{target}} \in \mathbb{R}^3$ denote the target position.
Let $T_i \le T_{\max}$ denote the terminal navigation step of the episode.
The UGV-to-target distance at step $t$ is defined as $d_{i,t}^{g}
=
\left\|
\mathbf{x}_{i,t}^{g}
-
\mathbf{x}_{i}^{\mathrm{target}}
\right\|_2.$ The success indicator $S_i \in \{0,1\}$ is defined as
\begin{equation}
S_i
=
\mathbb{I}
\left[
q_{i,T_i}=1
\;\land\;
d_{i,T_i}^{g}\le d_s
\right],
\end{equation}
where $q_{i,T_i}=1$ indicates that the collaborative navigation system outputs the termination signal at step $T_i$, and $d_s=5\,\mathrm{m}$ denotes the success distance threshold. Based on these episode-level definitions, we compute the following evaluation metrics over the $N$ episodes:

\begin{itemize}[leftmargin=1.5em, itemsep=2pt, topsep=3pt, parsep=0pt]

\item \textbf{Success Rate (SR).}
SR measures the fraction of successful episodes:
\begin{equation}
\mathrm{SR}
=
\frac{1}{N}
\sum_{i=1}^{N}
S_i.
\end{equation}

\item \textbf{Oracle Success Rate (OSR).}
OSR measures the fraction of episodes in which the UGV reaches within the
success distance threshold at any point along its trajectory:
\begin{equation}
\mathrm{OSR}
=
\frac{1}{N}
\sum_{i=1}^{N}
\mathbb{I}
\left[
\min_{0\le t\le T_i}
d_{i,t}^{g}
\le d_s
\right].
\end{equation}

\item \textbf{Success weighted by Path Length (SPL).}
Let $\ell_i$ and $p_i$ denote the shortest traversable path length and the
actual path length of the UGV, respectively. SPL is defined as
\begin{equation}
\mathrm{SPL}
=
\frac{1}{N}
\sum_{i=1}^{N}
S_i
\frac{\ell_i}{\max(\ell_i,p_i)}.
\end{equation}

\item \textbf{UGV Navigation Error (NE\textsubscript{UGV}).}
$\mathrm{NE}_{\mathrm{UGV}}$ measures the average 3D distance between the UGV
and the target at the terminal navigation step:
\begin{equation}
\mathrm{NE}_{\mathrm{UGV}}
=
\frac{1}{N}
\sum_{i=1}^{N}
d_{i,T_i}^{g}.
\end{equation}

\item \textbf{UAV Navigation Error (NE\textsubscript{UAV}).}
As a diagnostic metric for the UAV, $\mathrm{NE}_{\mathrm{UAV}}$ measures
the average horizontal distance between the UAV and the target at the terminal
navigation step. Let $\Pi_{xy}(\cdot)$ denote projection onto the horizontal
plane:
\begin{equation}
\mathrm{NE}_{\mathrm{UAV}}
=\frac{1}{N}
\sum_{i=1}^{N}
\left\|
\Pi_{xy}\!\left(\mathbf{x}_{i,T_i}^{a}\right)
-
\Pi_{xy}\!\left(\mathbf{x}_{i}^{\mathrm{target}}\right)
\right\|_2.
\end{equation}

\end{itemize}

\textbf{Evaluation Splits.}
The seen/unseen partition is defined at the environment level. The seen split
contains 17 environments used during training, while the unseen split contains
2 held-out environments. Seen evaluation uses new episodes from the training
environments with different start poses and relative target directions,
whereas unseen evaluation assesses generalization to held-out environments
and unseen target instances. To evaluate air--ground collaboration under different aerial visibility
conditions, we further categorize test episodes into \textit{aerial-visible}
and \textit{aerial-occluded} subsets based on whether the target is visible
from the aerial viewpoint.

\section{Benchmark Construction Details}
\label{supp:benchmark_construction}
\subsection{Instruction Construction}
The goal-oriented instruction generation component consists of five stages:
\textit{Target Selection}, \textit{Target Instance Description},
\textit{Environmental Context Description},
\textit{Coarse Direction Annotation}, and
\textit{Instruction Synthesis}. We detail each stage below.

\noindent\textbf{(1) Target Selection.}
We select target instances with distinctive visual attributes or contextual cues. Each candidate target is then reviewed by multiple annotators to verify that it can be consistently described and identified from language instructions.

\begin{figure*}[h]
    \centering
    \includegraphics[width=0.9\textwidth]{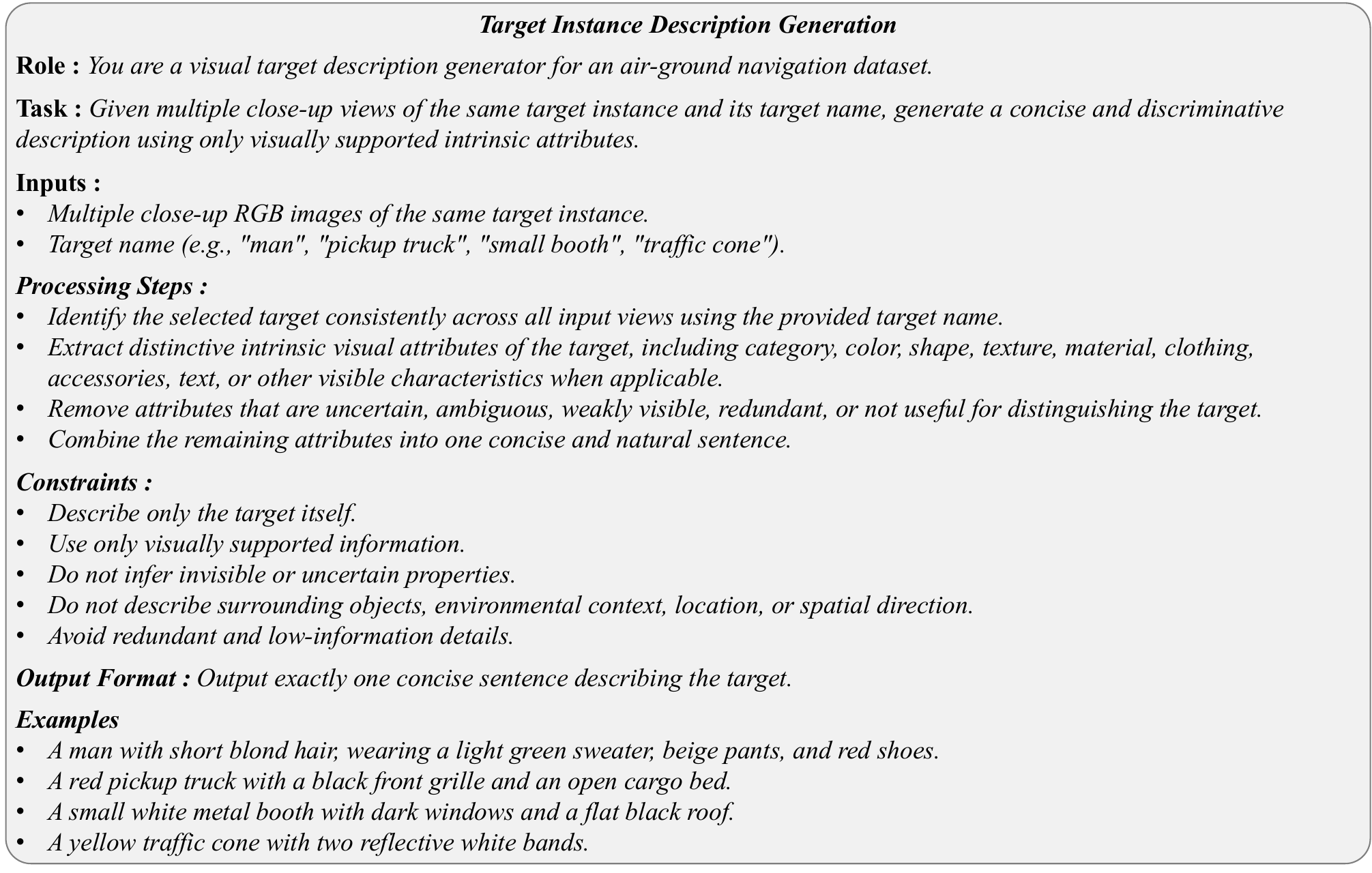}
    \caption{
    Prompt for target instance description.
    }
    \label{fig:target_description_generation}
\end{figure*}

\noindent\textbf{(2) Target Instance Description.}
For each selected target, we collect target-centric multi-view observations and use GPT-5.5 to generate a concise description of its semantic category and distinctive visual attributes. The description focuses on visually supported attributes of the target itself, while excluding surrounding environmental context. The corresponding prompt is shown in Fig.~\ref{fig:target_description_generation}.

\begin{figure*}[h]
    \centering
    \includegraphics[width=0.9\textwidth]{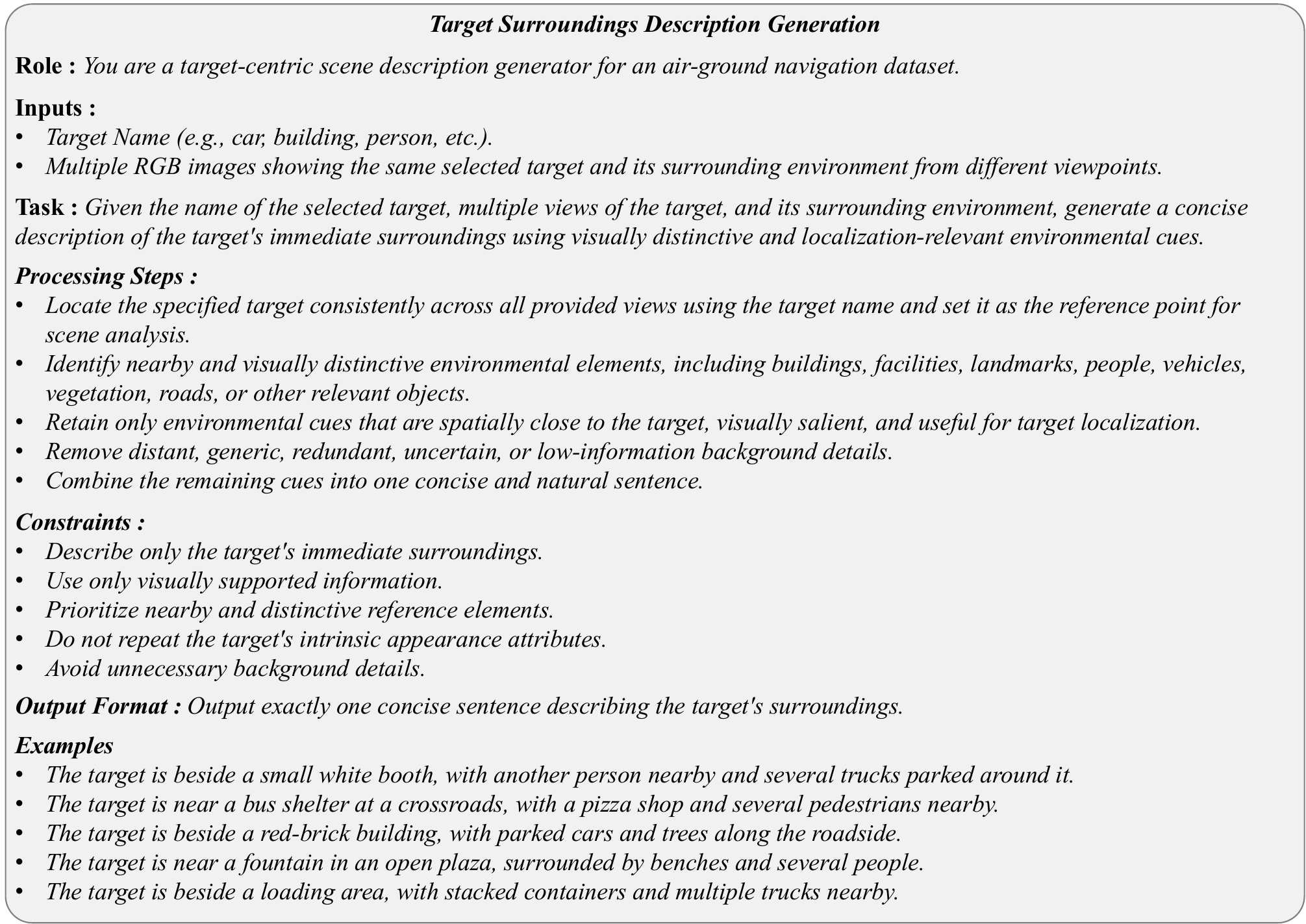}
    \caption{
    Prompt for environmental context description.
    }
    \label{fig:surroundings_description_generation}
\end{figure*}
\begin{figure*}[h]
    \centering
    \includegraphics[width=0.9\textwidth]{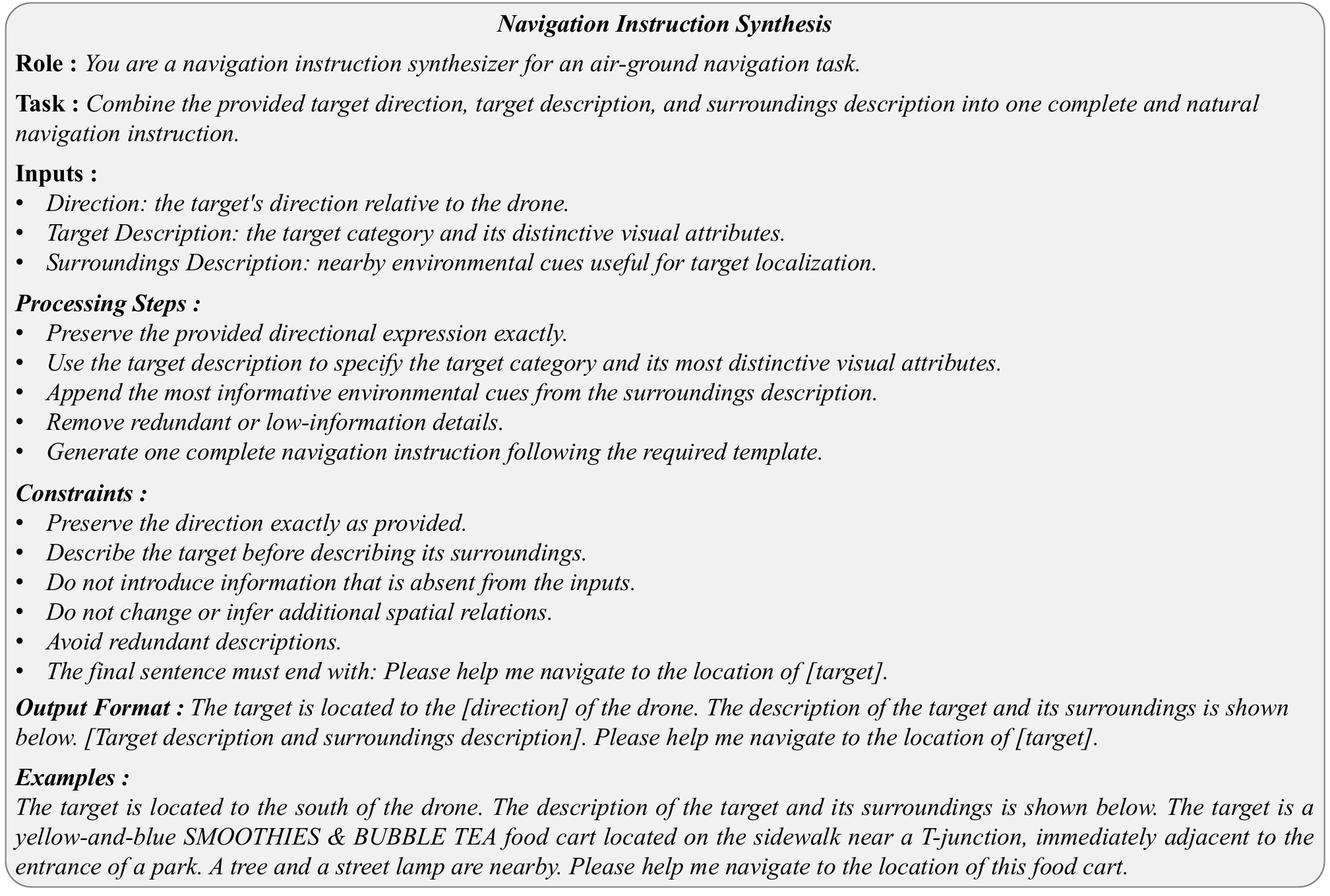}
    \caption{
    Prompt for goal-oriented instruction synthesis.
    }
    \label{fig:navigation_instruction_composition}
\end{figure*}

\noindent\textbf{(3) Target Surroundings Description.}
We further collect multi-view observations of the environment surrounding the target from multiple directions. GPT-5.5 summarizes nearby objects, structures,
and other localization-relevant environmental cues while avoiding redundant descriptions of the target itself. The corresponding prompt is shown in Fig.~\ref{fig:surroundings_description_generation}.

\noindent\textbf{(4) Coarse Direction Annotation.}
For each target, annotators select multiple episode start locations from different directions. Since valid ground start locations must satisfy local traversability constraints, we use human selection rather than automatic sampling to ensure high-quality episode initialization. Given the start and target coordinates, we automatically compute the coarse relative direction from the start location to the target and quantize it into eight compass directions: north, northeast, east, southeast, south, southwest, west, and northwest.

\noindent\textbf{(5) Navigation Instruction Synthesis.}
GPT-5.5 integrates the coarse relative direction, target description, and environmental context into the goal-oriented navigation instruction, without prescribing intermediate actions. The corresponding prompt is shown in Fig.~\ref{fig:navigation_instruction_composition}.

\begin{figure*}[h]
    \centering
    \includegraphics[width=\textwidth]{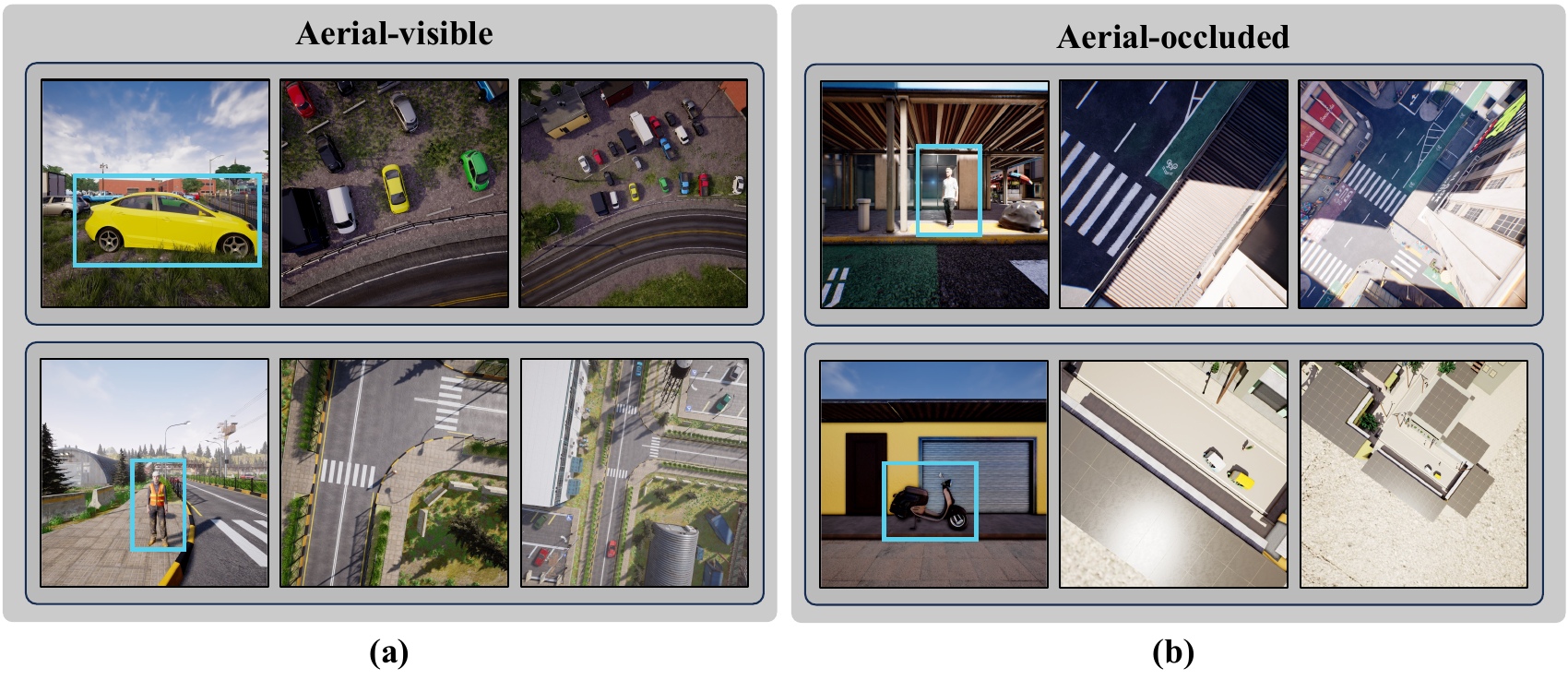}
    \caption{
    Representative aerial-visible and aerial-occluded targets.}
    \label{fig:aerial_visibility_examples}
\end{figure*}
\subsection{Aerial-Visibility Annotation.}
To evaluate collaborative navigation under different aerial target-visibility conditions, we partition all target instances in the test sets into \textit{aerial-visible} and \textit{aerial-occluded} subsets. Since current vision-language models remain unreliable for fine-grained spatial occlusion reasoning, we manually annotate each target instance to ensure annotation quality. A target is labeled \textit{aerial-visible} if it is directly observable from the top-down aerial viewpoint, and \textit{aerial-occluded}  if it is hidden from this viewpoint by roofs, overhangs, tree canopies, or other scene elements.  Representative examples are shown in Fig.~\ref{fig:aerial_visibility_examples}.

\subsection{Human Quality Assurance.}
We adopt a multi-stage human verification process to ensure the quality of the constructed episodes. Each candidate episode is reviewed for consistency across the target instance, goal-oriented instruction, air--ground reference trajectories, and aerial-visibility annotation. Reviewers verify that the target is uniquely identifiable, the instruction accurately describes the intended target and its surrounding context, and both aerial and ground trajectories are valid and executable. Episodes with correctable issues are manually revised and then cross-checked by a different annotator before being retained. Samples with ambiguous targets, invalid trajectories, or annotation errors that cannot be reliably resolved are discarded. For uncertain cases or disagreements between reviewers, an additional annotator examines the episode and determines the final annotation. After episode-level verification, all retained samples undergo a final dataset-level review to identify remaining inconsistencies across annotations and trajectories. Of the 14,697 initially constructed episodes, 6,375 were accepted without modification, 3,906 were retained after correction and re-verification, and 4,416 were discarded, resulting in the final 10,281 episodes.

\section{Additional Dataset Statistics}
\label{supp:data_splits}
\begin{figure*}[h]
    \centering
    \includegraphics[width=\textwidth]{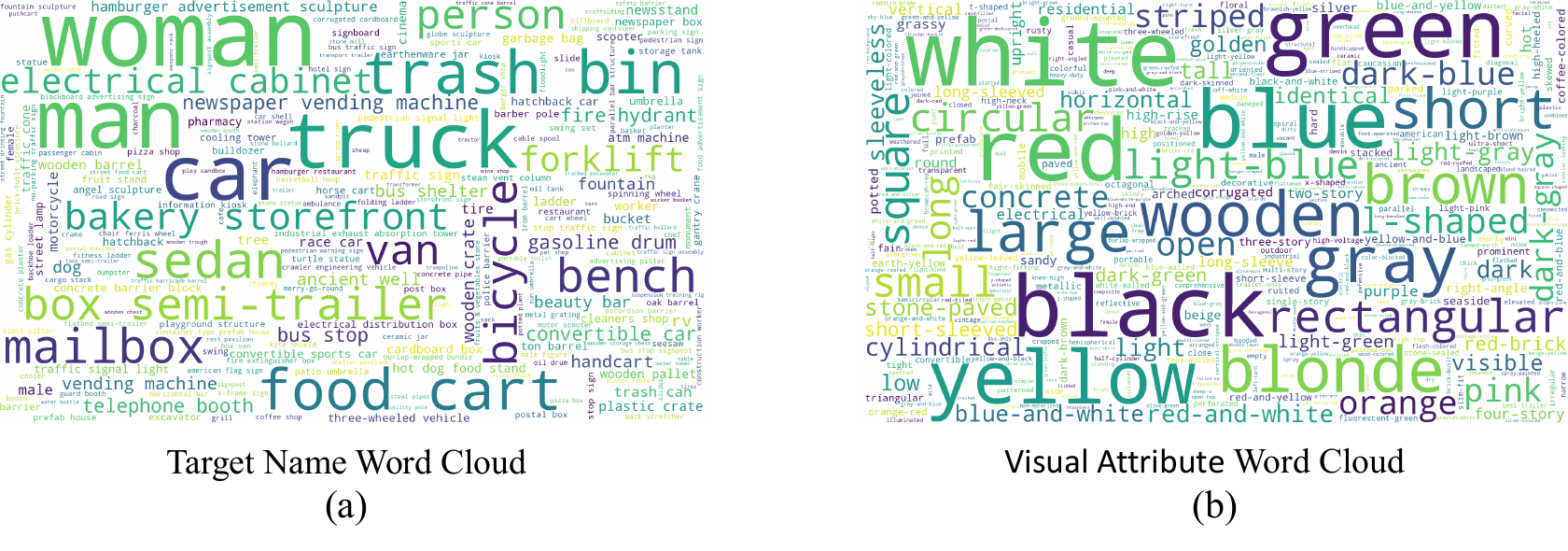}
    \caption{
    Additional statistics of AirGroundVLN.
    (a) Target name word cloud.
    (b) Visual attribute word cloud.
    }
    \label{fig:additional_dataset_statistics}
\end{figure*}

\noindent\textbf{Target and Visual Attribute Statistics.}
The word cloud in Fig.~\ref{fig:additional_dataset_statistics} (a) further
highlights the semantic diversity of target entities, ranging from common
vehicles and pedestrians to infrastructure-related objects such as fire
hydrants, electrical cabinets, and traffic signal lights. Meanwhile,
Fig.~\ref{fig:additional_dataset_statistics} (b) shows that target
descriptions contain diverse fine-grained visual attributes, including
color, shape, material, texture, and scale. Together, these statistics characterize the category-level semantics and
instance-level appearance cues represented in AirGroundVLN.

\noindent\textbf{Aerial Visibility Distribution.} The test episodes are categorized into \textit{aerial-visible} and
\textit{aerial-occluded} cases, with 77.3\% and 22.7\% of episodes falling into
each category, respectively.

\section{Method Details}
\label{supp:method}

\subsection{Training Objective.}
\label{supp:training objective}
For platform $v\in\{g,a\}$, let $Z_t^v\in\mathbb{R}^{H\times W}$ denote the
predicted waypoint logits. The supervision target
$Y_t^v\in\mathbb{R}^{H\times W}$ is constructed as a normalized Gaussian heatmap centered at the ground-truth future waypoint of platform $v$. The predicted waypoint distribution is $P_t^v=\operatorname{Softmax}(Z_t^v).$ We combine distribution-level supervision with an auxiliary coordinate-level
objective:
\begin{equation}
\mathcal{L}_v
=
D_{\mathrm{KL}}
\left(
Y_t^v \parallel P_t^v
\right)
+
\lambda_c
\operatorname{SmoothL1}
\left(
\mu(P_t^v),
\mu(Y_t^v)
\right),
\label{eq:platform_loss}
\end{equation}
where $\mu(\cdot)$ denotes the differentiable expected spatial coordinate under a probability map. Let $\mathbf{r}_{ij}\in\mathbb{R}^2$ denote the spatial
coordinate associated with grid cell $(i,j)$. The spatial expectation is
computed as $\mu(P)=\sum_{i=1}^{H}\sum_{j=1}^{W}P_{ij}\mathbf{r}_{ij}.$ The KL divergence supervises the predicted waypoint distribution with the
Gaussian supervision distribution, while the coordinate-level term further constrains its expected spatial coordinate. The aerial and ground
objectives are jointly optimized as
\begin{equation}
\mathcal{L}_{\mathrm{total}}
=
\mathcal{L}_g
+
\lambda_a\mathcal{L}_a,
\label{eq:total_loss}
\end{equation}
where $\lambda_c$ controls the coordinate-level supervision and $\lambda_a$
balances the aerial objective relative to the ground objective. 

\subsection{Target Approach and Termination}
\label{supp:target_approach}

Once a target candidate is confirmed, its 3D position
$\hat{\mathbf{x}}^{\mathrm{target}}$ is estimated from synchronized
RGB-D observations in the same coordinate frame as the UGV. Since the
detected target may not lie on the traversable surface, only the horizontal
coordinates $\hat{\mathbf{x}}^{\mathrm{target}}_{xy}$ are used as the approach
goal. The UGV then executes a local planner to approach this estimated target
position~\citep{apex}. No additional VLM-based waypoint prediction is performed during this
final approach stage. The approach radius is set to
$d_{\mathrm{stop}}=1.5\,\mathrm{m}$, measured in the horizontal plane, and
the maximum approach budget is $T_{\mathrm{app}}=20$ iterations. The approach
terminates when the UGV is within $d_{\mathrm{stop}}$ of
$\hat{\mathbf{x}}^{\mathrm{target}}_{xy}$ or when the execution budget is
exhausted. The stopping radius is a controller termination condition and is
distinct from the benchmark success criterion, which evaluates the final UGV
pose against the ground-truth target position.

\section{Implementation Details}
\label{supp:experiments}
AG-CoNAV is built upon the pretrained Qwen3.5-2B~\citep{Qwen35}
vision-language backbone. We keep the pretrained backbone weights
frozen and train LoRA adapters~\citep{LoRA}. We use
AdamW~\citep{AdamW} with an initial learning rate of
$2\times10^{-5}$ and a cosine learning-rate decay schedule. For waypoint supervision, we set the coordinate-level loss weight to
$\lambda_c=10.0$ and the aerial loss weight to
$\lambda_a=0.5$. The SACM memory bank stores up to
$M=16$ historical air--ground entries.
The maximum navigation budget is set to $T_{\max}=200$ steps. During navigation, the UGV uses only a front-facing camera. After executing each step, it performs a 360$^\circ$ sweep using 6 camera angles for target verification.
Once a target candidate is confirmed, the UGV enters the final
approach stage, for which we set the stopping radius to
$d_{\mathrm{stop}}=1.5\,\mathrm{m}$ and the maximum number
of approach steps to $T_{\mathrm{app}}=20$. All methods use the same target verification and final approach
procedure to ensure a fair comparison. 

\begin{figure*}[t]
\centering
\includegraphics[width=\textwidth]{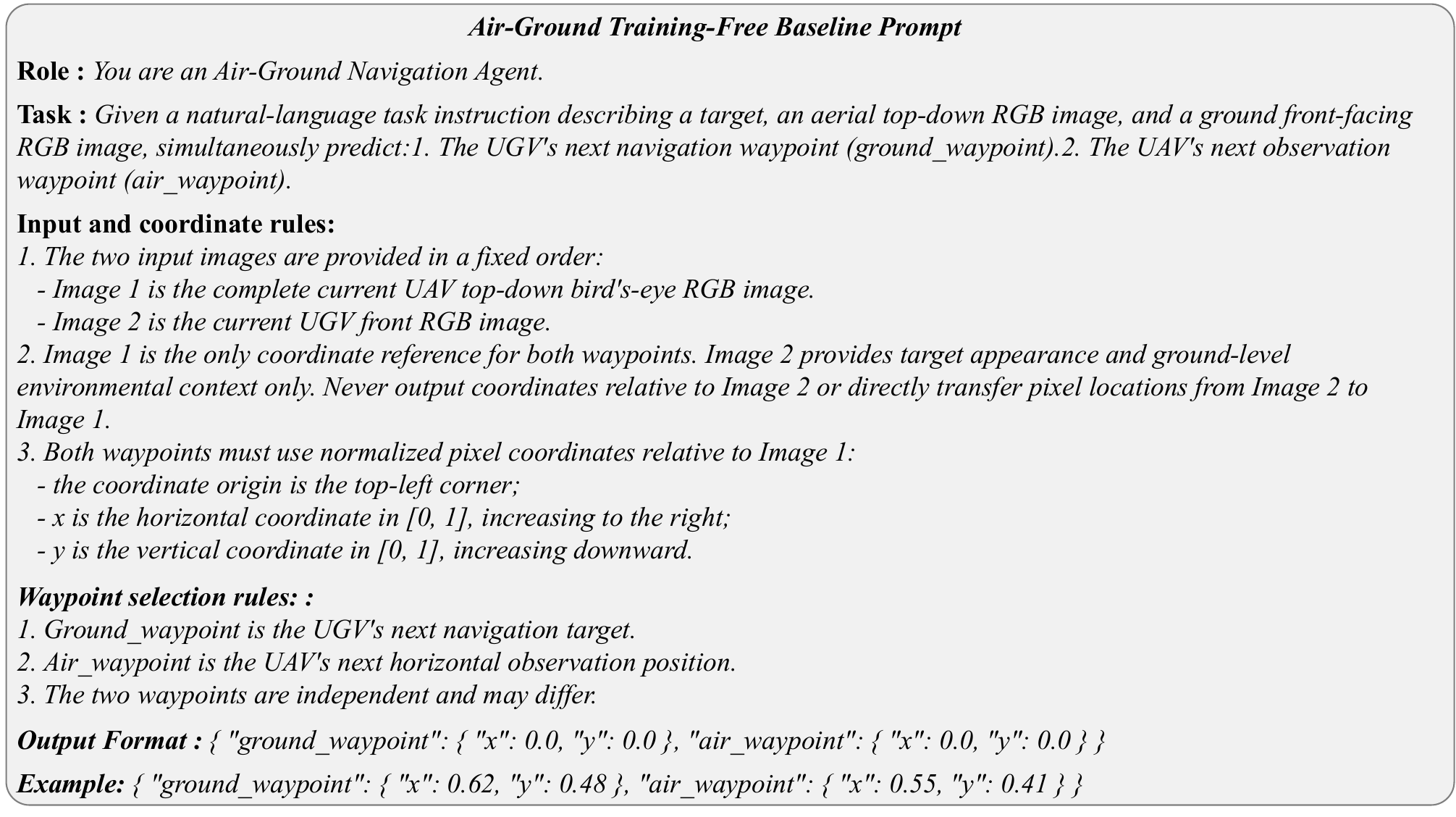}
\caption{
Air--ground waypoint generation prompt.
}
\label{fig:training_free_waypoint_prompt}
\end{figure*}

\begin{figure*}[t]
\centering
\includegraphics[width=\textwidth]{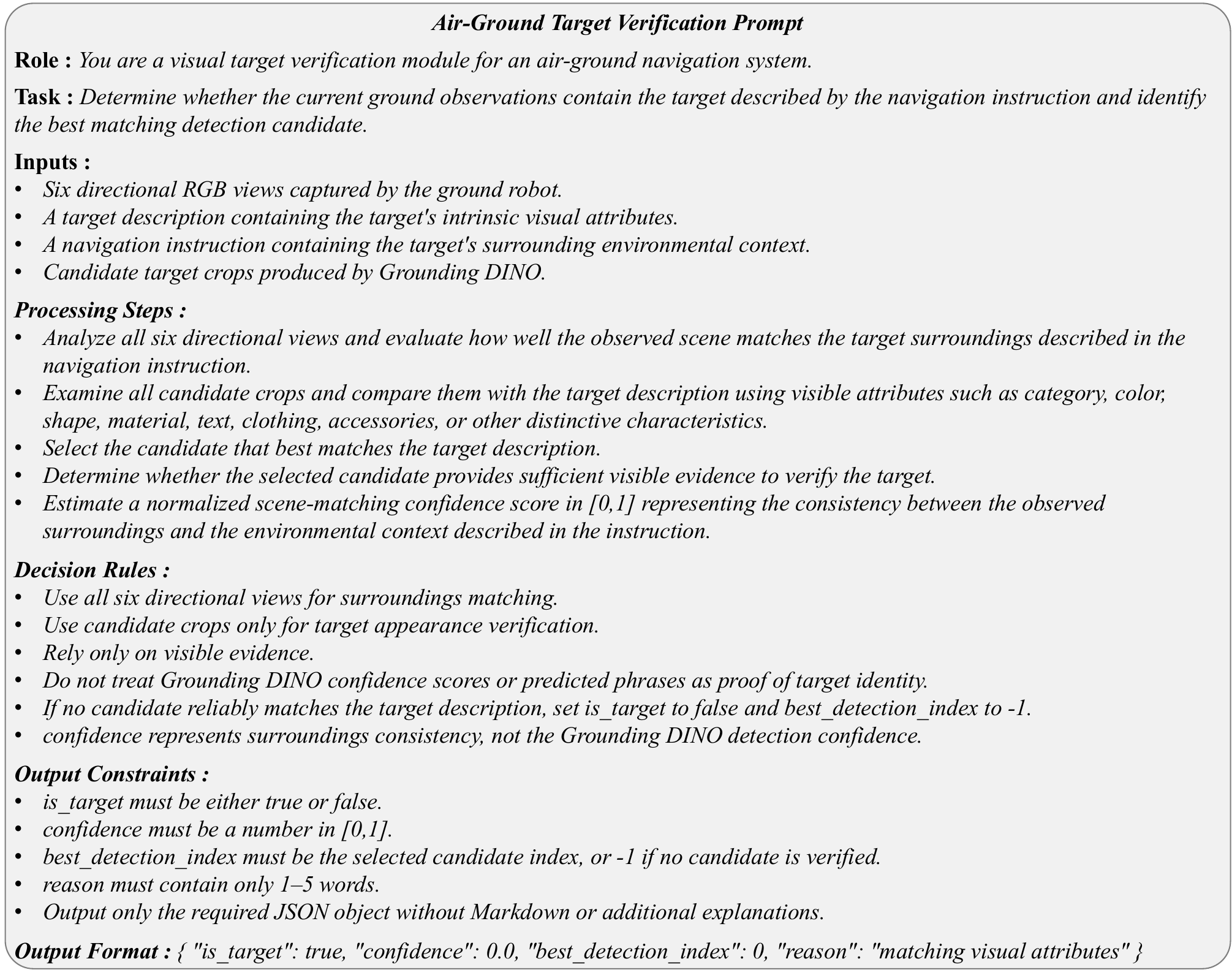}
\caption{
Target verification prompt.
}
\label{fig:target_verification_prompt}
\end{figure*}

\noindent\textbf{Training-Free Baseline Setup.}
For the VLM-based training-free baselines, the model receives the goal-oriented instruction together with the current ground and aerial BEV observations and predicts the next UGV and UAV waypoints. The overall navigation procedure is summarized in Algorithm~\ref{alg:airground_trainingfree}. The waypoint generation model $\mathcal{M}_{\mathrm{VLM}}$ uses the prompt shown in Fig.~\ref{fig:training_free_waypoint_prompt} to jointly predict the next UGV navigation waypoint $\hat{w}_t^g$ and UAV observation waypoint $\hat{w}_t^a$ from the instruction and current air--ground observations. The target verifier $\mathcal{V}$ uses the prompt shown in
Fig.~\ref{fig:target_verification_prompt} to determine whether the current observations match the instructed target by jointly considering candidate appearance attributes and surrounding environmental context. It outputs a confidence score $s_t$ used for termination.

\begin{algorithm}[h]
\small
\caption{Training-Free Air--Ground Navigation}
\label{alg:airground_trainingfree}
\begin{algorithmic}[1]
\Require instruction $I$, initial poses $(P_0^g,P_0^a)$,
maximum navigation budget $T_{\max}$,
verification threshold $\tau$,
stopping radius $d_{\mathrm{stop}}$,
maximum approach steps $T_{\mathrm{app}}$

\State Initialize the UGV and UAV at $(P_0^g,P_0^a)$ and set $t\gets0$.
\While{$t<T_{\max}$}
    \State Acquire the current ground and aerial BEV observations
    $(o_t^g,o_t^a)$.
    
    \State Verify the target using appearance and surrounding context matching: $(b_t,s_t)\gets\mathcal{V}(I,o_t^g,o_t^a)$, where $b_t$ indicates whether the target is verified and $s_t$ denotes the target-context matching confidence.

    \If{$b_t \land (s_t\geq\tau)$}
    \State Estimate the confirmed target position
    $\hat{\mathbf{x}}^{\mathrm{target}}$ from RGB-D observations.
    \State Execute final approach toward
    $\hat{\mathbf{x}}^{\mathrm{target}}$ with stopping radius
    $d_{\mathrm{stop}}$ and step limit $T_{\mathrm{app}}$.
    \State Terminate the episode.
    \State \textbf{break}
    \EndIf

    \State Predict air--ground waypoints
    $(\hat{w}_t^g,\hat{w}_t^a)
    \gets
    \mathcal{M}_{\mathrm{VLM}}(I,o_t^g,o_t^a)$.

    \State Generate navigation trajectories
    $\hat{u}_t^g\gets\mathcal{P}_g(P_t^g,\hat{w}_t^g)$ and
    $\hat{u}_t^a\gets\mathcal{P}_a(P_t^a,\hat{w}_t^a)$.

    \State Execute $(\hat{u}_t^g,\hat{u}_t^a)$ and update
    $(P_{t+1}^g,P_{t+1}^a)$.
    \State $t\gets t+1$.
\EndWhile
\end{algorithmic}
\end{algorithm}

\section{Qualitative Navigation Analysis}
\label{supp:qualitative}
\begin{figure*}[h]
\centering
\includegraphics[width=\textwidth]{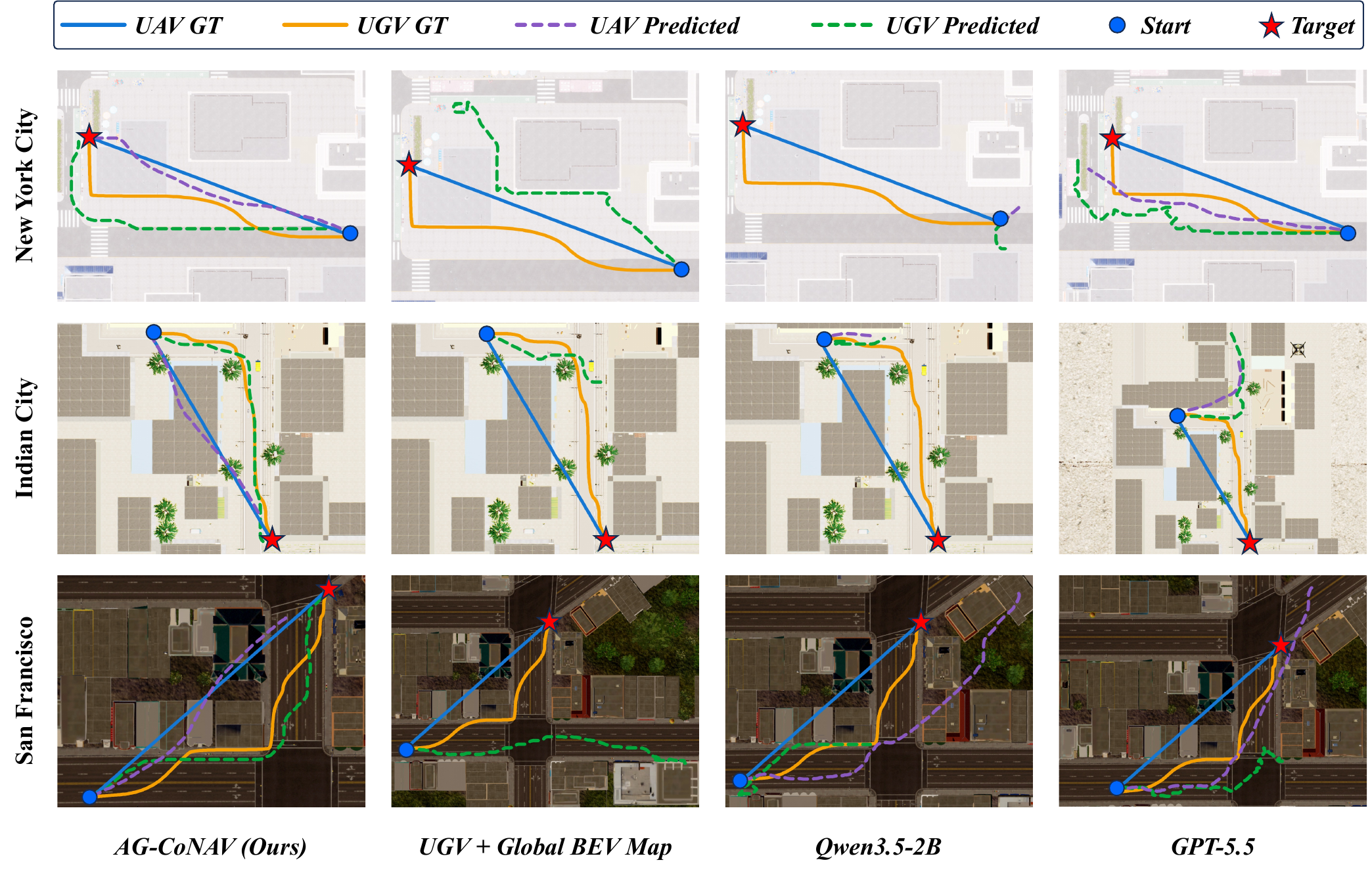}
\caption{
\textbf{Bird's-eye-view trajectory comparisons on AirGroundVLN.}
Joint UAV and UGV trajectories from four methods (Qwen3.5-2B, GPT-5.5, \textit{UGV + Global BEV Map}, and AG-CoNAV) are visualized across three representative episodes alongside ground-truth (GT) paths.
}
\label{fig:trajectory_bev}
\end{figure*}

\begin{figure*}[h]
\centering
\includegraphics[width=\textwidth]{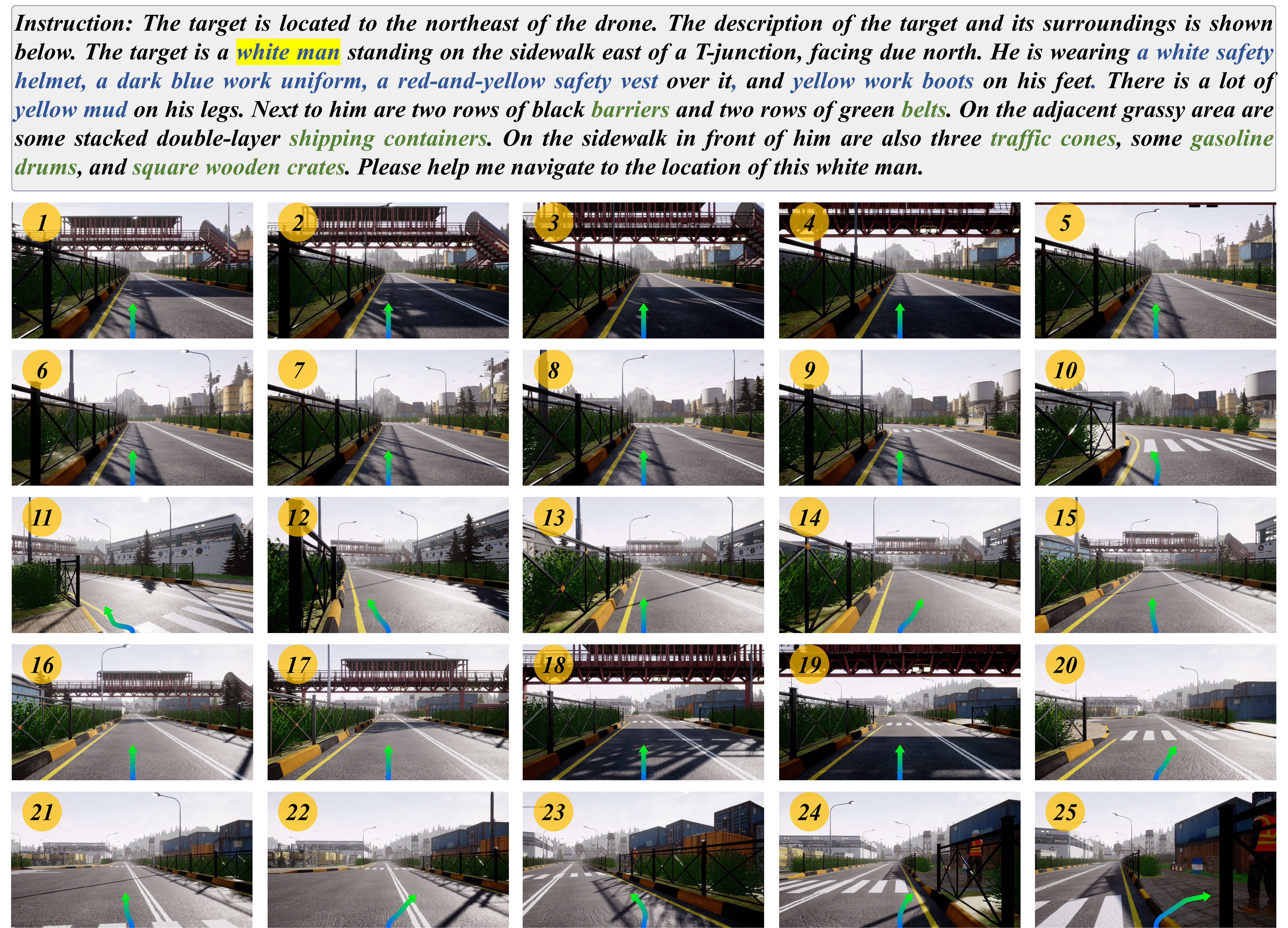}
\caption{
\textbf{Qualitative navigation trajectory visualization on AirGroundVLN.}
Navigation trajectories toward language-specified targets are visualized in
\textit{\textbf{seen}} environments with \textit{\textbf{visible}} aerial observations.
}
\label{fig:trajectory_seen}
\end{figure*}

\begin{figure*}[h]
\centering
\includegraphics[width=\textwidth]{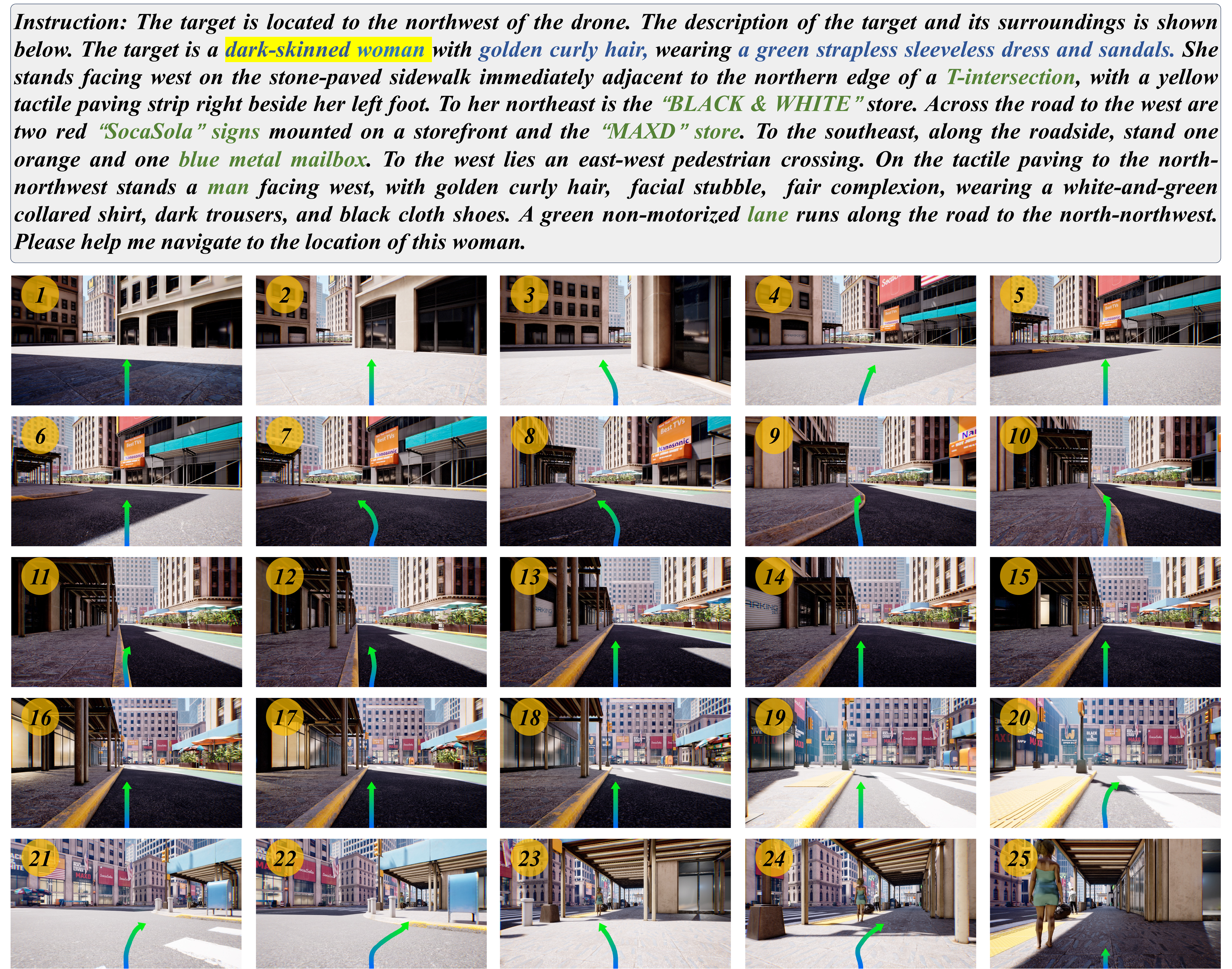}
\caption{
\textbf{Qualitative navigation trajectory visualization on AirGroundVLN.}
Navigation trajectories toward language-specified targets are visualized in
\textit{\textbf{unseen}} environments with \textit{\textbf{occluded}} aerial observations.
}
\label{fig:trajectory_unseen}
\end{figure*}

To demonstrate the navigation behavior of different approaches, we present qualitative comparisons across multiple scenarios in AirGroundVLN. First, Fig.~\ref{fig:trajectory_bev} provides bird's-eye-view (BEV) trajectory visualizations across three distinct navigation episodes, including \textit{\textbf{two episodes from unseen environments}}, comparing our proposed AG-CoNAV against representative training-free baselines (Qwen3.5-2B, GPT-5.5) and a training-based baseline (\textit{UGV + Global BEV Map}), alongside ground-truth (GT) paths. As shown in Fig.~\ref{fig:trajectory_bev}, AG-CoNAV consistently reaches the target locations across all three scenarios with reasonable path execution. In contrast, the zero-shot baselines (Qwen3.5-2B and GPT-5.5) often select incorrect directions at crucial intersections due to the difficulty of long-range spatial reasoning without task-specific training. Meanwhile, the \textit{UGV + Global BEV Map} baseline, constrained by a static global map and a lack of fine-grained local details, frequently mistargets inaccurate goals or fails during execution. Furthermore, Fig.~\ref{fig:trajectory_seen} and Fig.~\ref{fig:trajectory_unseen} detail step-by-step navigation visualizations in \textit{seen-visible} and \textit{unseen-occluded} environments, respectively. Notably, AG-CoNAV maintains accurate target localization and executable path planning even in \textit{\textbf{unseen}} environments with \textit{\textbf{aerial occlusions}}.

\end{document}